\documentclass[journal]{IEEEtran}

\newcommand{\forarxiv}{0}

\usepackage[sort,compress]{cite}
\usepackage{wrapfig}
\usepackage[dvipsnames]{xcolor}
\usepackage{tikz}
\usepackage{bbm}
\usepackage{enumerate}
\colorlet{NextBlue}{red!25!green!50!blue!75}

\newcommand{\carrlfiguredir}{figures/journal}

\usetikzlibrary{shapes,arrows,fit,backgrounds}
\tikzstyle{block} = [draw, rectangle, text width=2cm, text centered, minimum height=1.2cm, node distance=3cm]
\tikzstyle{container} = [draw, rectangle, inner sep=0.3cm, fill=NextBlue!20, minimum height=3cm]
\def\bottom#1#2{\hbox{\vbox to #1{\vfill\hbox{#2}}}}
\usetikzlibrary{positioning,calc}

\tikzset{
  mybackground/.style={execute at end picture={
      \begin{scope}[on background layer]
        \node[font=\bf] at (current bounding box.north){\bottom{1cm} #1};
        \end{scope}
    }},
}

\tikzset{
  zigzag/.style={
    to path={
      coordinate (m) at ($(\tikztostart)!.5!(\tikztotarget)$)
      coordinate (m1) at ($(m)!1mm!110:(\tikztostart)$)
      coordinate (m2) at ($(m)!1mm!110:(\tikztotarget)$)
      plot[rounded corners=1mm] coordinates{ (\tikztostart) (m1) (m2) (\tikztotarget)}
    }
  },
}

\usepackage{graphicx} 
\usepackage{amsfonts}
\usepackage{amsmath,soul}
\usepackage{amssymb} 
\usepackage[hidelinks]{hyperref}
\usepackage[capitalize]{cleveref} 
\usepackage{dsfont} 
\usepackage[font=small,skip=0pt]{subcaption} 
\usepackage{balance}
\usepackage[font=small]{caption}
\usepackage{bm} 
\usepackage{wrapfig}

\if \forarxiv 1
\usepackage[final]{changes} 
\else
\usepackage[authormarkuptext=name,authormarkupposition=left]{changes}
\fi
\definechangesauthor[name={J.~H.}, color={blue}]{jh}
\definechangesauthor[name={M.~E.}, color={red}]{me}
\definechangesauthor[name={B.~L.}, color={green}]{bl}
\definechangesauthor[name={G.~H.}, color={orange}]{gh}
\DeclareMathOperator*{\argmin}{\arg\!\min}
\DeclareMathOperator*{\argmax}{\arg\!\max}
\newcommand{\citep}{\cite}
\newcommand{\cmmnt}[1]{\ignorespaces}

\usepackage[acronym]{glossaries}
\makeglossaries
\newacronym{rl}{RL}{Reinforcement Learning}
\newacronym{sl}{SL}{Supervised Learning}
\newacronym{dnn}{DNN}{Deep Neural Network}
\newacronym{dnns}{DNN}{Deep Neural Networks}
\newacronym{lstm}{LSTM}{Long Short-Term Memory}
\newacronym{carrl}{CARRL}{Certified Adversarial Robustness for Deep Reinforcement Learning}
\newacronym{orca}{ORCA}{Optimal Reciprocal Collision Avoidance}
\newacronym{relu}{ReLU}{Rectified Linear Unit}
\newacronym{dqn}{DQN}{Deep Q-Network}
\newacronym{lp}{LP}{Linear Programming}
\newacronym{smt}{SMT}{Satisfiability Modulo Theory}
\newacronym{fgst}{FGST}{Fast Gradient Sign method with Targeting}
\newacronym{crown}{CROWN}{Algorithm from~\cite{Weng_2018b}}
\newacronym{ga3c}{GA3C}{Hybrid GPU/CPU Asynchronous Advantage Actor-Critic}
\newacronym{cadrl}{CADRL}{Collision Avoidance with Deep Reinforcement Learning}
\newacronym{ga3ccadrl}{GA3C-CADRL}{\acrfull{ga3c} \acrshort{cadrl}}

\usepackage{amsthm}
\newtheorem{theorem}{Theorem}[section]

\newtheorem{definition}[theorem]{Definition}

\newcommand{\papertitle}{Certifiable Robustness to Adversarial State Uncertainty in Deep Reinforcement Learning}

\begin{document}
\title{\papertitle}

\author{
\if \forarxiv 1
Michael~Everett*,
Bj\"{o}rn~L\"{u}tjens*,
and~Jonathan~P.~How
\else
Michael~Everett*,~\IEEEmembership{Student Member,~IEEE,}
Bj\"{o}rn~L\"{u}tjens*,~\IEEEmembership{Student Member,~IEEE,}
and~Jonathan~P.~How,~\IEEEmembership{Fellow,~IEEE}
\fi
\thanks{Authors are with the Aerospace Controls Laboratory, Massachusetts Institute of Technology, Cambridge,
MA, 02139 USA e-mail: mfe@mit.edu}
\thanks{Preliminary version of paper presented at Conference on Robot Learning (CoRL) on Nov. 1, 2019~\cite{Lutjens19_CORL}. \if \forarxiv 0 {This manuscript submitted Mar. 28, 2020 with revised versions submitted Aug. 21, 2020 and Jan. 22, 2021, and accepted Jan. 28, 2021.}\fi}
\thanks{* indicates equal contributions}
}

\if \forarxiv 1
\markboth{(in review)}%
{Everett \MakeLowercase{\textit{et al.}}: \papertitle}
\else
\markboth{Accepted to IEEE Transactions on Neural Networks and Learning Systems}%
{Everett \MakeLowercase{\textit{et al.}}: \papertitle}
\fi

\maketitle

\begin{abstract}
Deep Neural Network-based systems are now the state-of-the-art in many robotics tasks, but their application in safety-critical domains remains dangerous without formal guarantees on network robustness.
Small perturbations to sensor inputs (from noise or adversarial examples) are often enough to change network-based decisions, which was recently shown to cause an autonomous vehicle to swerve into another lane.
In light of these dangers, numerous algorithms have been developed as defensive mechanisms from these adversarial inputs, some of which provide formal robustness guarantees or certificates.
This work leverages research on certified adversarial robustness to develop an online \textit{certifiably robust} for deep reinforcement learning algorithms.
The proposed defense computes guaranteed lower bounds on state-action values during execution to identify and choose a robust action under a worst-case deviation in input space due to possible adversaries or noise.
Moreover, the resulting policy comes with a \textit{certificate} of solution quality, even though the true state and optimal action are unknown to the certifier due to the perturbations.
The approach is demonstrated on a Deep Q-Network policy and is shown to increase robustness to noise and adversaries in pedestrian collision avoidance scenarios, a classic control task, and Atari Pong.
This work extends~\cite{Lutjens19_CORL} with new performance guarantees, extensions to other RL algorithms, expanded results aggregated across more scenarios, an extension into scenarios with adversarial behavior, comparisons with a more computationally expensive method, and visualizations that provide intuition about the robustness algorithm.
\end{abstract}

\begin{IEEEkeywords}
Adversarial Attacks, Reinforcement Learning, Collision Avoidance, Robustness Verification
\end{IEEEkeywords}

\IEEEpeerreviewmaketitle


\section{Introduction} \label{sec:intro}

\IEEEPARstart{D}{eep} \acrfull*{rl} algorithms have achieved impressive success on robotic manipulation~\citep{Gu_2017} and robot navigation in pedestrian crowds~\citep{fan2019getting,Everett_2018}.
Many of these systems utilize black-box predictions from \acrfull*{dnns} to achieve state-of-the-art performance in prediction and planning tasks.
However, the lack of formal robustness guarantees for \acrshort*{dnn}s currently limits their application in safety-critical domains, such as collision avoidance.
In particular, even subtle perturbations to the input, known as \textit{adversarial examples}, can lead to incorrect (but highly-confident) decisions by \acrshort*{dnn}s~\citep{Szegedy_2014, Akhtar_2018, Yuan_2019}.
Furthermore, several recent works have demonstrated the danger of adversarial examples in real-world situations~\citep{Kurakin_2017b, Sharif_2016}, including causing an autonomous vehicle to swerve into another lane~\citep{Tencent_2019}.
The work in this paper not only addresses deep \acrshort*{rl} algorithms' lack of robustness against network input uncertainties, but it also provides formal guarantees in the form of certificates on the solution quality.

While there are many techniques for synthesis of \textit{empirically} robust deep RL policies~\citep{mandlekar2017adversarially,Rajeswaran_2017,Muratore_2018,Pinto_2017,morimoto2005robust}, it remains difficult to synthesize a \textit{provably} robust neural network.
Instead, we leverage ideas from a related area that provides a guarantee on how sensitive a trained network's output is to input perturbations for each nominal input~\citep{Ehlers_2017, Katz_2017, Huang_2017b,Lomuscio_2017,Tjeng_2019,Gehr_2018}.
A promising recent set of methods makes such formal robustness analysis tractable by relaxing the nonlinear constraints associated with network activations~\citep{Wong_2018,wang2018efficient,dvijotham2018training,singh2018fast,singh2019abstract,Weng_2018,Weng_2018b}, unified in~\citep{salman2019convex}.
These relaxed methods were previously applied on computer vision or other \acrfull*{sl} tasks.

This work extends the tools for efficient formal neural network robustness analysis (e.g., \citep{Weng_2018,singh2018fast,Wong_2018}) to deep \acrshort*{rl} tasks.
In \acrshort*{rl}, techniques designed for \acrshort*{sl} robustness analysis would simply allow a nominal action to be \textit{flagged} as non-robust if the minimum input perturbation exceeds a robustness threshold (e.g., the system's known level of uncertainty) -- these techniques would not reason about alternative actions.

Hence, we instead focus on the \textit{robust decision-making problem}: given a known bound on the input perturbation, what is the best action to take?
This aligns with the requirement that an agent \textit{must} select an action at each step of an \acrshort*{rl} problem.
Our approach uses robust optimization to consider worst-case observational uncertainties and provides certificates on solution quality, making it \textit{certifiably robust}.
The proposed algorithm is called \textbf{C}ertified \textbf{A}dversarial \textbf{R}obustness for Deep \textbf{RL} (\textbf{CARRL}).

\begin{figure}[tp]
    \vspace*{-.15in}
    \centering
    \includegraphics[width=0.35\textwidth]{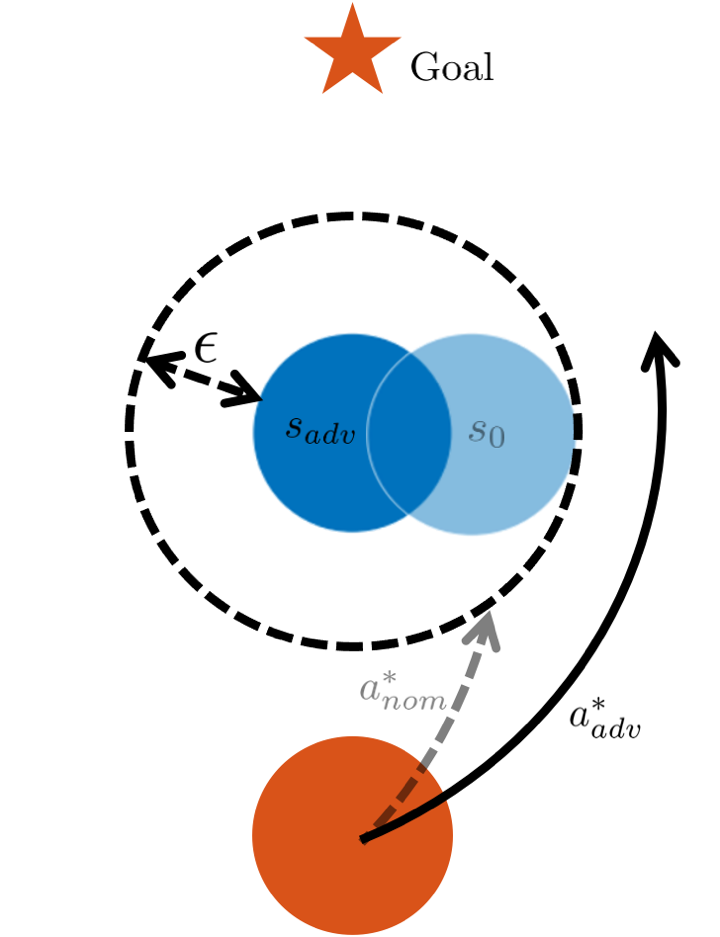}
    \caption[Intuition on Certified Adversarial Robustness]{Intuition. An adversary distorts the true position, $\bm{s}_0$, of a dynamic obstacle (blue) into an adversarial observation, $\bm{s}_\text{adv}$. The agent (orange) only sees the adversarial input, so nominal \acrshort*{rl} policies would take $a^*_\text{nom}$ to reach the goal quickly, but would then collide with the true obstacle, $\bm{s}_0$. The proposed defensive strategy considers that $\bm{s}_0$ could be anywhere inside the $\bm{\epsilon}$-ball around $\bm{s}_\text{adv}$, and selects the action, ${a^*_\text{adv}}$, with the best, worst-case outcome as calculated by a guaranteed lower bound on the value network output, which cautiously avoids the obstacle while reaching the goal. Note this is different from simply inflating the obstacle radius, since the action values contain information about environment dynamics, e.g., blue agent's cooperativeness.}
    \label{fig:intuition_vfy_rl}
\end{figure}

As a motivating example, consider the collision avoidance setting in~\cref{fig:intuition_vfy_rl}, in which an adversary perturbs an agent's (orange) observation of an obstacle (blue).
An agent following a nominal/standard deep \acrshort*{rl} policy would observe $\bm{s}_\text{adv}$ and select an action, $a^*_\text{nom}$, that collides with the obstacle's true position, $\bm{s}_0$, thinking that the space is unoccupied.
Our proposed approach assumes a worst-case deviation of the observed input, $\bm{s}_\text{adv}$, bounded by $\bm{\epsilon}$, and takes the robust-optimal action, $a^*_\text{adv}$, under that perturbation, to safely avoid the true obstacle.
Robustness analysis algorithms often assume $\bm{\epsilon}$ is a scalar, which makes sense for image inputs (all pixels have same scale, e.g., $0{-}255$ intensity).
A key challenge in direct application to \acrshort*{rl} tasks is that the observation vector (network input) could have elements with substantially different scales (e.g., position, angle, joint torques) and associated measurement uncertainties, motivating our extension with vector $\bm{\epsilon}$.

This work contributes (i) the first formulation of \textit{certifiably robust deep \acrshort*{rl}}, which uses robust optimization to consider worst-case state perturbations and provides certificates on solution quality, (ii) an extension of tools for efficient robustness analysis to handle variable scale inputs common in \acrshort*{rl}, and (iii) demonstrations of increased robustness to adversaries and sensor noise on cartpole and a pedestrian collision avoidance simulation.

We extend~\cite{Lutjens19_CORL} with new performance guarantees (\cref{sec:approach:guarantees}), a formulation for policies with reduced sensitivity (\cref{sec:approach:reduced_sensitivity_policy}), algorithmic extensions to policy-based RL (\cref{sec:approach:policy_based_rl}), expanded results aggregated across more scenarios (\cref{sec:results:cartpole,sec:results:collision_avoidance}), an extension of the algorithm into scenarios with adversarial behavior (\cref{sec:results:behavioral_adversary}), comparisons with a more computationally expensive method (\cref{sec:results:carrl_vs_lp}), and visualizations that provide intuition about the robustness algorithm (\cref{sec:results:intuition_on_bounds}).

\section{Related work} \label{sec:related_work}

The lack of robustness of \acrshort*{dnn}s to real-world uncertainties~\citep{Szegedy_2014} has motivated the study of adversarial attacks (i.e., worst-case uncertainty realizations) in many learning tasks.
This section summarizes adversarial attack and defense models in deep \acrshort*{rl} (see~\citep{ilahi2020challenges} for a thorough survey) and describes methods for formally quantifying \acrshort*{dnn} robustness.

\subsection{Adversarial Attacks in Deep RL}

An adversary can act against an \acrshort*{rl} agent by influencing (or exploiting a weakness in) the observation or transition models of the environment.
\subsubsection*{Observation model}
Many of the techniques for attacking \acrshort*{sl} networks through small image perturbations~\citep{Goodfellow_2015} could be used to attack image-based deep \acrshort*{rl} policies.
Recent works show how to specifically craft adversarial attacks (in the input image space) against a \acrfull*{dqn} in \acrshort*{rl}~\citep{Huang_2017,behzadan2017vulnerability,yang2020enhanced}.
Another work applies both adversarial observation and transition perturbations \citep{mandlekar2017adversarially}.
\subsubsection*{Transition model}
Several approaches attack an \acrshort*{rl} agent by changing parameters of the physics simulator, like friction coefficient or center of gravity, between episodes~\citep{Rajeswaran_2017,Muratore_2018}.
Other approaches change the transition model between steps of episodes, for instance by applying disturbance forces~\citep{Pinto_2017,morimoto2005robust}, or by adding a second agent that is competing against the ego agent~\citep{Uther_1997}.
In~\citep{gleave2019adversarial}, the second agent unexpectedly learns to visually distract the ego agent rather than exerting forces and essentially becomes an observation model adversary.
Thus, adversarial behavior of a second agent (the topic of \textit{multiagent games}~\citep{littman1994markov}) introduces complexities beyond the scope of this work (except a brief discussion in \cref{sec:results:behavioral_adversary}), which focuses on robustness to adversarial observations.

\subsection{Empirical Defenses to Adversarial Attacks}

Across deep learning in general, the goal of most existing robustness methods is to make neural networks' outputs less sensitive to small input perturbations.

\subsubsection*{Supervised Learning}

First introduced for \acrshort*{sl} tasks, adversarial training or retraining augments the training dataset with adversaries~\citep{Kurakin_2017, Madry_2018, Kos_2017, Mirman_2019} to increase robustness during testing (empirically).
Other works increase robustness through distilling networks~\citep{Papernot_2016} or comparing the output of model ensembles~\citep{Tramer_2018}.
Rather than modifying the network training procedure, another type of approach \textit{detects} adversarial examples through comparing the input with a binary filtered transformation of the input~\citep{Xu_2018}.
Although these approaches show impressive empirical success, they are often ineffective against more sophisticated adversarial attacks~\citep{Carlini_2017, He_2017, Athalye_2018, Uesato_2018}.

\begin{figure*}[tp]
    \vspace*{-.15in}
    \centering
    \includegraphics[width=\textwidth, clip, trim=0 320 0 0]{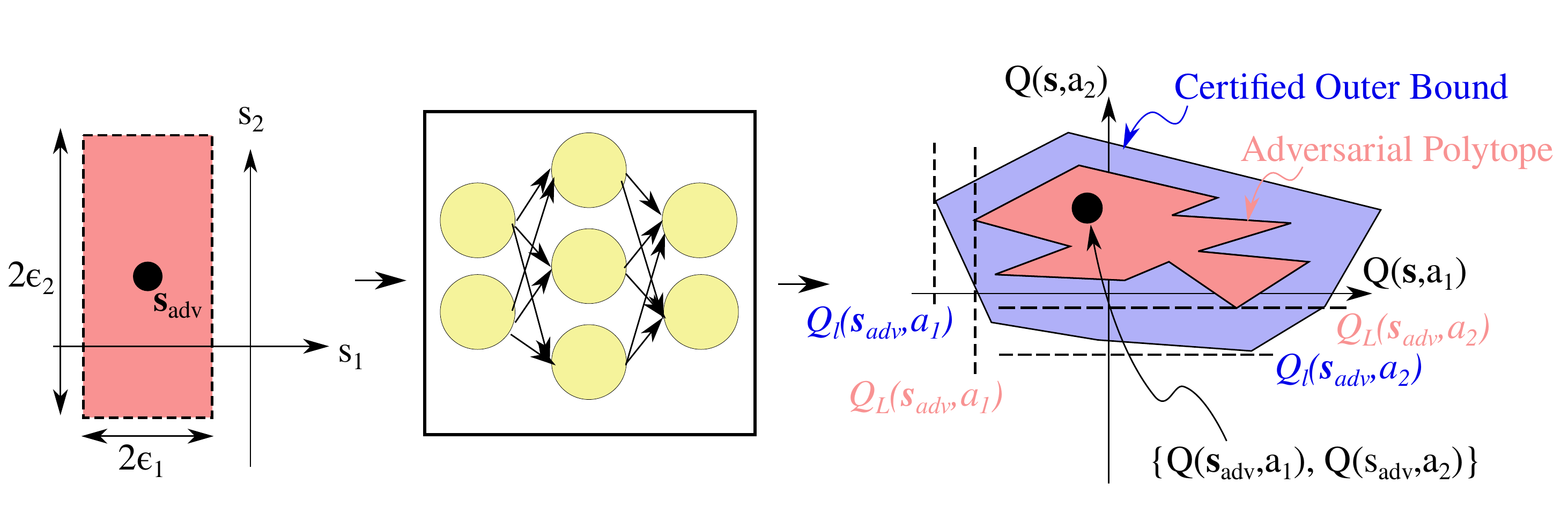}
    \caption[State Uncertainty Propagated Through \acrlong{dqn}]{
    State Uncertainty Propagated Through \acrlong{dqn}.
    The red region (left) represents bounded state uncertainty (\mbox{an $\ell_{\infty}\ \bm{\epsilon}$-ball}) around the observed state, $\bm{s}_\text{adv}$.
    A neural network maps this set of possible inputs to a polytope (red) of possible outputs (Q-values in \acrshort*{rl}).
    This work's extension of~\cite{Weng_2018} provides an outer bound (blue) on that polytope.
    This work then modifies the \acrshort*{rl} action-selection rule by considering lower bounds, $Q_l$, on the blue region for each action.
    In this 2-state, 2-action example, our algorithm would select action 2, since $Q_l(\bm{s}_\text{adv},a_2)>Q_l(\bm{s}_\text{adv},a_1)$, i.e. the worst possible outcome from action 2 is better than the worst possible outcome from action 1, given an $\bm{\epsilon}$-ball uncertainty and a pre-trained DQN.
    }
    \label{fig:state_uncertainty_cartoon}
\end{figure*}

\subsubsection*{Deep \acrshort*{rl}}
Many ideas from \acrshort*{sl} were transferred over to deep \acrshort*{rl} to provide empirical defenses to adversaries (e.g., training in adversarial environments~\citep{mandlekar2017adversarially,Rajeswaran_2017,Muratore_2018,Pinto_2017,morimoto2005robust}, using model ensembles~\citep{Rajeswaran_2017,Muratore_2018}).
Moreover, because adversarial observation perturbations are a form of measurement uncertainty, there are close connections between Safe \acrshort*{rl}~\citep{Garcia_2015} and adversarial robustness.
Many Safe \acrshort*{rl} (also called risk-sensitive \acrshort*{rl}) algorithms optimize for the reward under \textit{worst-case} assumptions of environment stochasticity, rather than optimizing for the expected reward~\citep{Heger_1994,Tamar_2015,Geibel_2006}.
The resulting policies are more risk-sensitive (i.e., robust to stochastic deviations in the input space, such as sensor noise), but could still fail on algorithmically crafted adversarial examples.
In other words, modifying the \acrshort*{rl} training process to directly synthesize a \textit{provably} robust neural network remains challenging.

Instead, this work adds a defense layer on top of an already trained \acrshort*{dqn}.
We provide guarantees on the robustified policy's solution quality by leveraging formal robustness analysis methods that propagate known \acrshort*{dnn} input bounds to guaranteed output bounds.

\subsection{Formal Robustness Methods}\label{sec:lit_review_formal_robustness}

Although synthesis of a provably robust neural network is difficult, there is a closely related body of work that can provide other formal guarantees on network robustness -- namely, a guarantee on how sensitive a trained network's output is to input perturbations.
The corresponding mathematical machinery to propagate an input set through a neural network allows for solving various robustness analysis problems, such as the calculation of \textit{reachable sets}, \textit{minimal adversarial examples}, or \textit{robustness verification}.

For example, \textit{exact methods} can be used to find tight bounds on the maximum network output deviation, given a nominal input and bounded input perturbation.
These methods rely on \acrfull*{smt}~\citep{Ehlers_2017, Katz_2017, Huang_2017b}, \acrshort*{lp}/mixed-integer \acrshort*{lp} solvers~\citep{Lomuscio_2017, Tjeng_2019}, or zonotopes~\citep{Gehr_2018}, to propagate constraints on the input space through to the output space (exactly).
The difficulty in this propagation arises through \acrshort*{dnn}s with \acrfull*{relu} or other nonlinear activation functions -- in fact, the problem of finding the exact output bounds is NP-complete (as shown by \citep{Katz_2017,Weng_2018}), making real-time implementations infeasible for many robotics tasks.

The formal robustness analysis approaches can be visualized in~\cref{fig:state_uncertainty_cartoon} in terms of a 2-state, 2-action deep \acrshort*{rl} problem.
For a given state uncertainty set (red, left), the exact methods reason over the exact adversarial polytope (red, right), i.e., the image of the state uncertainty set through the network.

To enable scalable analysis, many researchers simultaneously proposed various relaxations of the NP-complete formulation: \citep{salman2019convex} provides a unifying framework of the \textit{convex relaxations} that we briefly summarize here.
A convex relaxation of the nonlinear activations enables the use of a standard \acrshort*{lp} solver to compute convex, guaranteed outer bounds, $Q_{LP}$, on the adversarial polytope (cf. Problem $\mathcal{C}$ in~\citep{salman2019convex}).
Even though this \acrshort*{lp} can be solved more efficiently than the exact problem with nonlinear activations, solving the relaxed \acrshort*{lp} is still computationally intensive.

Fortunately, the \acrshort*{lp} can be solved greedily, and thus even more efficiently, by using just \textit{one} linear upper bound and \textit{one} linear lower bound for each nonlinear layer.
Several seemingly different approaches (e.g., solving the dual problem~\citep{Wong_2018,wang2018efficient,dvijotham2018training}, using zonotopes/polyhedra as in~\cite{singh2018fast,singh2019abstract}, using linear outer bounds~\cite{Weng_2018,Weng_2018b}) were shown to lead to very similar or identical network bounds in~\cite{salman2019convex}.
These greedy solutions provide over-approximations (see blue region in~\cref{fig:state_uncertainty_cartoon}) on the \acrshort*{lp} solution and thus, also on the adversarial polytope.
One of these methods, Fast-Lin~\citep{Weng_2018} (which this work extends), can verify an image classification (MNIST) network's sensitivity to perturbations for a given image in $<200ms$.
As new approaches appear in the literature, such as~\citep{cohen2019certified}, it will be straightforward to ``upgrade'' this work's use of~\citep{Weng_2018} for calculating lower bounds.

The analysis techniques described here in~\cref{sec:lit_review_formal_robustness} are often formulated to solve the \textit{robustness verification problem}: determine whether the calculated set of possible network outputs crosses a decision hyperplane (a line of slope 1 through the origin, in this example) -- if it does cross, the classifier is deemed ``not robust'' to the input uncertainty.
Our approach instead solves the \textit{robust decision-making problem}: determine the best action considering worst-case outcomes, $Q_l(\bm{s}_\text{adv}, a_1), Q_l(\bm{s}_\text{adv}, a_2)$, denoted by the dashed lines in \cref{fig:state_uncertainty_cartoon}.

\subsection{Certified Robustness in RL vs. SL}

The inference phase of reinforcement learning (RL) and supervised learning (SL) are similar in that the policy chooses an action/class given some current input. However, there are at least three key differences between the proposed certified robustness for RL and existing certification methods for SL:

\begin{enumerate}
    \item Existing methods for SL certification do not actually change the decision, they simply provide an additional piece of information (``is this decision sensitive to the set of possible inputs? yes/no/unsure'') alongside the nominal decision.
    Most SL certification papers do not discuss what to do if the decision \textit{is} sensitive, which we identify as a key technical gap since in
    RL it is not obvious what to do with a sensitivity flag, as an action still needs to be taken.
    Thus, instead of just returning the nominal decision plus a sensitivity flag, the method in our paper makes a robust decision that considers the full set of inputs.
    \item In SL, there is a ``correct'' output for each input (the true class label).
    In RL, the correct action is less clear because the actions need to be compared/evaluated in terms of their impact on the system.
    Furthermore, in RL, some actions could be much worse than others, whereas in SL typically all ``wrong'' outputs are penalized the same.
    Our method proposes using the Q-value encoded in a DNN as a way of comparing actions under state uncertainty.
    \item In SL, a wrong decision incurs an immediate cost, but that decision does not impact the future inputs/costs.
    In RL, actions affect the system state, which \textit{does} influence future inputs/rewards.
    This paper uses the value function to encode that information about the future, which we believe is an important first step towards considering feedback loops in the certification process.
\end{enumerate}

To summarize, despite SL and RL inference appearing similarly, there are several key differences in the problem contexts that motivate a fundamentally different certification framework for RL.

\section{Background} \label{sec:background}

\subsection{Preliminaries}

In \acrshort*{rl} problems\footnote{This work considers problems with a continuous state space and discrete action space.}, the state-action value (or, ``Q-value'')
\begin{equation}
Q(\bm{s},a)=\mathop{\mathbb{E}}_{\mathbf{s}'\sim P}\left[\sum_{t=0}^{T}{\gamma^t r(t)}|\mathbf{s}(t{=}0)=\bm{s},\mathrm{a}(t{=}0)=a\right], \nonumber
\end{equation}
expresses the expected accumulation of future reward, $r$, discounted by $\gamma$, received by starting in state $\bm{s}\in\mathbb{R}^n$ and taking one of $d$ discrete actions, $a\in \{a_0, a_1, \ldots, a_{d-1}\}=\mathcal{A}$, with each next state $\mathbf{s}'$ sampled from the (unknown) transition model, $P$.
Furthermore, when we refer to Q-values in this work, we really mean a \acrshort*{dnn} approximation of the Q-values.

Let $\bm{\epsilon}\in\mathbb{R}^n_{\geq 0}$ be the maximum element-wise deviation of the state vector, and let $1\leq p \leq \infty$ parameterize the $\ell_p$-norm.
We define the set of states within this deviation as the $\bm{\epsilon}$-Ball,
\begin{align}
\mathcal{B}_p(\bm{s}_0, \bm{\epsilon}) &= \{\bm{s} : \lim_{\bm{\epsilon}' \to \bm{\epsilon}^+} \lvert\lvert (\bm{s} - \bm{s}_0) \oslash \bm{\epsilon}' \rvert\rvert_p \leq 1\},
\label{eq:eps_ball}
\end{align}
where $\oslash$ denotes element-wise division, and the $\mathrm{lim}$ is only needed to handle the case where $\exists i\ \epsilon_i=0$ (e.g., when the adversary is not allowed to perturb some component of the state, or the agent knows some component of the state vector with zero uncertainty).

An $\bm{\epsilon}$-Ball is illustrated in~\cref{fig:eps_ball} for the case of $n=2$, highlighting that different elements of the state, $s_i$, might have different perturbation limits, $\epsilon_i$, and that the choice of $\ell_p$-norm affects the shape of the ball.
The $\ell_p$-norm is defined as ${\lvert\lvert \bm{x} \rvert\rvert_p = (\lvert x_1 \lvert^p + \ldots + \lvert x_n \lvert^p)^{1/p}\ \mathrm{for}\ \bm{x}\in\mathbb{R}^{n}, 1\leq p \leq \infty}$.

\newcommand\lonecolor{CarnationPink}
\newcommand\ltwocolor{CornflowerBlue}
\newcommand\linfcolor{Dandelion}
\newcommand\scolor{Violet}
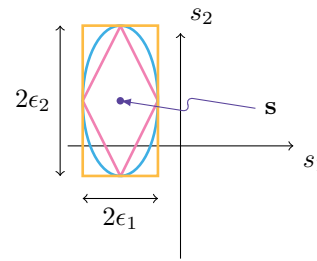
\begin{figure}
    \centering
    \begin{tikzpicture}
        \newcommand\epsone{0.5}
        \newcommand\epstwo{1.0}
        \newcommand\offsetx{-0.8}
        \newcommand\offsety{0.6}
        \newcommand\normlinethickness{1}
        \newcommand\epsilonlabeloffset{0.3}
        \draw[->] (-1.5,0) -- (1.5,0) node[anchor=north west]{$s_1$};
        \draw[->] (0,-1.5) -- (0,1.5) node[anchor=south west]{$s_2$};
        \coordinate (e1) at (\epsone+\offsetx,\offsety);
        \coordinate (e2) at (0+\offsetx,\epstwo+\offsety);
        \coordinate (e3) at (-\epsone+\offsetx,0+\offsety);
        \coordinate (e4) at (0+\offsetx,-\epstwo+\offsety);
        \coordinate (s0) at (\offsetx, \offsety);
        \node (s0label) at (1.2, 0.5) {$\mathbf{s}$};

        \draw[line width=\normlinethickness, \lonecolor] (e1) -- (e2)+(0.0,-0.01) -- (e3) -- (e4)+(0.0,-0.01) --(e1);
        
        \draw[line width=\normlinethickness, \ltwocolor] (s0) ellipse ({\epsone} and {\epstwo});
        
        \draw[line width=\normlinethickness, \linfcolor] (-\epsone+\offsetx,-\epstwo+\offsety) rectangle (\epsone+\offsetx,\epstwo+\offsety);

        \node[circle,inner sep=1pt,fill=\scolor] at (s0) {};
        \draw[\scolor,-latex] (s0label.west) to[zigzag] (s0);

        \draw[<->] (-\epsone+\offsetx,\offsety-\epstwo-\epsilonlabeloffset) -- (\epsone+\offsetx,\offsety-\epstwo-\epsilonlabeloffset) node[midway, below]{$2\epsilon_1$};
        \draw[<->] (-\epsone+\offsetx-\epsilonlabeloffset,\offsety-\epstwo) -- (-\epsone+\offsetx-\epsilonlabeloffset,\offsety+\epstwo) node[midway, left]{$2\epsilon_2$};

    \end{tikzpicture}
    \caption{
    Illustration of $\bm{\epsilon}$-Ball, $\mathcal{B}_p(\bm{s},\bm{\epsilon})$, for $n=2$.
    Let $\bm{\epsilon}=[\epsilon_1, \epsilon_2]$.
    Depending on the choice of $\ell_p$ norm (e.g., \textcolor{\lonecolor}{\boldmath{$\ell_1$}}, \textcolor{\ltwocolor}{\boldmath{$\ell_2$}}, and \textcolor{\linfcolor}{\boldmath{$\ell_\infty$}}),
    $\mathcal{B}_p(\bm{s},\bm{\epsilon})$ is the set of points inside the corresponding colored outline.
    The adversary can perturb nominal observation $\bm{s}$ to any point inside $\mathcal{B}_p(\bm{s},\bm{\epsilon})$.
    The values of $\{n,\bm{\epsilon},p\}$ are application-specific choices and the components of $\bm{\epsilon}$ need not be equal.
    }
    \label{fig:eps_ball}
\end{figure}

\subsection{Robustness Analysis}

This work aims to find the action that maximizes state-action value under a worst-case perturbation of the observation by sensor noise or an adversary.
This section explains how to efficiently obtain a lower bound on the \acrshort*{dnn}-predicted $Q$, given a bounded perturbation in the state space from the true state.
The derivation is based on~\citep{Weng_2018}, re-formulated for \acrshort*{rl}.

The adversary perturbs the true state, $\bm{s}_0$, to another state, ${\bm{s}_\text{adv} \in \mathcal{B}_{p_\text{adv}}(\bm{s}_{0}, \bm{\epsilon}_\text{adv})}$, within the $\bm{\epsilon}_\text{adv}$-ball.
The ego agent only observes the perturbed state, ${\bm{s}_\text{adv}}$.
As displayed in~\cref{fig:state_uncertainty_cartoon}, let the worst-case state-action value, $Q_L$, for a given action, $a_j$, be 
\begin{align}
Q_L(\bm{s}_\text{adv},a_j) &= \min_{\bm{s} \in \mathcal{B}_{p_\text{adv}}(\bm{s}_\text{adv}, \bm{\epsilon}_\text{adv})} Q(\bm{s}, a_j), \label{eq:qlj_lower_bnd}
\end{align}
for all states $\bm{s}$ inside the $\bm{\epsilon}_\text{adv}$-Ball around the observation, $\bm{s}_\text{adv}$.

The goal of the analysis is to compute a guaranteed lower bound, $Q_l(\bm{s},a_j)$, on the minimum state-action value, that is, $Q_l(\bm{s},a_j) \leq Q_L(\bm{s}, a_j)$.
The key idea is to pass interval bounds\footnote{Element-wise $\pm\bm{\epsilon}_\text{adv}$ can cause overly conservative categorization of \acrshort*{relu}s for $p<\infty$. $p$ is accounted for later in the Algorithm in \cref{eq:qlj_full}.} $[\bm{l}^{(0)}, \bm{u}^{(0)}] = [\bm{s}_\text{adv} - \bm{\epsilon}_\text{adv}, \bm{s}_\text{adv} + \bm{\epsilon}_\text{adv}]$ from the \acrshort*{dnn}'s input layer to the output layer, where $\bm{l}^{(k)}$ and $\bm{u}^{(k)}$ denote the lower and upper bounds of the pre-activation term, $\bm{z}^{(k)}$, i.e., $l_i^{(k)}{\leq}z_i^{(k)}{\leq}u_i^{(k)}\forall i\in{1,...,u_k}$, in the $k$-th layer with $u_k$ units of an $m$-layer \acrshort*{dnn}.
When passing interval bounds through a \acrshort*{relu} activation\footnote{Although this work considers \acrshort*{dnn}s with \acrshort*{relu} activations, the formulation could be extended to general activation functions via more recent algorithms~\citep{Weng_2018b}.}, $\sigma(\cdot)$, the upper and lower pre-\acrshort*{relu} bounds of each element could either both be positive $(l_r^{(k)}, u_r^{(k)} > 0)$, negative $(l_r^{(k)}, u_r^{(k)} < 0)$, or positive and negative $(l_r^{(k)}<0, u_r^{(k)}>0)$, in which the \acrshort*{relu} status is called \textit{active, inactive} or \textit{undecided}, respectively.
In the active and inactive case, bounds are passed directly to the next layer.
In the undecided case, the output of the \acrshort*{relu} is bounded linearly above and below:
\begin{equation}
\begin{aligned}
\sigma(z_{r}^{(k)})_{|l_{r}^{(k)}, u_{r}^{(k)}}= 
    \begin{cases}
        [z_{r}^{(k)}, z_{r}^{(k)}] &\\\quad\quad\text{if $l_{r}^{(k)}, u_{r}^{(k)} > 0$, ``active''} \vspace{0.1in}\\
        [0,0] &\\\quad\quad\text{if $l_{r}^{(k)}, u_{r}^{(k)} < 0$, ``inactive''} \vspace{0.1in} \\ 
        [\frac{u_{r}^{(k)}}{u_{r}^{(k)}-l_{r}^{(k)}}z_{r}^{(k)}, \frac{u_{r}^{(k)}}{u_{r}^{(k)}-l_{r}^{(k)}}(z_{r}^{(k)}-l_{r}^{(k)})]&\\\quad\quad\text{if $l_{r}^{(k)}{<}0, u_{r}^{(k)}{>}0$, ``undecided'',}
    \end{cases}
\label{eq:sigma_lk_uk}
\end{aligned}
\end{equation}
for each element, indexed by $r$, in the $k$-th layer.

The identity matrix, $D$, is introduced as the \acrshort*{relu} status  matrix, $H$ as the lower/upper bounding factor, $W$ as the weight matrix, $b$ as the bias in layer $(k)$ with $r$ or $j$ as indices, and the pre-\acrshort*{relu}-activation, $z_r^{(k)}$, is replaced with $W_{r,:}^{(k)}\bm{s} + b_r^{(k)}$. The \acrshort*{relu} bounding is then rewritten as 
\begin{equation}\label{eq:bounds_on_sigma}
\begin{split}
& \quad D_{r,r}^{(k)}(W_{r,j}^{(k)}s_j + b_r^{(k)}) \\
\leq &\quad \sigma(W_{r,j}^{(k)}s_j + b_r^{(k)}) \\
\leq &\quad D_{r,r}^{(k)}(W_{r,j}^{(k)}s_j + b_r^{(k)} - H_{r,j}^{(k)}),
\end{split}
\end{equation}
where
\begin{align} 
D_{r,r}^{(k)} &=
    \begin{cases}
        1 &\text{if $l_r^{(k)}, u_r^{(k)}{>}0$;} \\
        0 &\text{if $l_r^{(k)}, u_r^{(k)}{<}0$,} \\
        \frac{u_r^{(k)}}{u_r^{(k)} - l_r^{(k)}} & \text{if $l_r^{(k)}{<}0, u_r^{(k)}{>}0$;}
    \end{cases}\\
H_{r,j}^{(k)} &=
    \begin{cases}
        l_r^{(k)} &\text{if $l_r^{(k)}{<}0, u_r^{(k)}{>}0, A_{j,r}^{(k)}{<}0$;}\\
        0 &\text{otherwise.}
    \end{cases}
\label{eq:D_H_defns}
\end{align}
Using these \acrshort*{relu} relaxations, a guaranteed lower bound of the state-action value for a single state $\bm{s}\in\mathcal{B}_p(\bm{s}_\text{adv}, \bm{\epsilon})$ (based on \cite{Weng_2018}) is:
\begin{equation}
\begin{aligned}
\bar{Q}_l(\bm{s}, a_j) = A_{j,:}^{(0)}\bm{s} + b_j^{(m)} + \sum_{k=1}^{m-1}{A_{j,:}^{(k)}(\bm{b}^{(k)}-H_{:,j}^{(k)})},
\label{eq:q_certified_lower_bound_closed_form}
\end{aligned}
\end{equation}
where the matrix $A$ contains the network weights and \acrshort*{relu} activation, recursively for all layers: $A^{(k-1)}=A^{(k)}W^{(k)}D^{(k-1)}$, with identity in the final layer: $A^{(m)}=\mathds{1}$.

Unlike the exact \acrshort*{dnn} output, the bound on the relaxed \acrshort*{dnn}'s output in~\cref{eq:q_certified_lower_bound_closed_form} can be minimized across an $\bm{\epsilon}$-ball in closed form (as described in~\cref{sec:approach:variable_e}), which is a key piece of this work's real-time, robust decision-making framework.


\section{Approach} \label{sec:approach}
This work develops an add-on, certifiably robust defense for existing Deep \acrshort*{rl} algorithms to ensure robustness against sensor noise or adversarial examples during test time.

\subsection{System architecture}
\begin{figure*}[t]
  \centering
  \begin{tikzpicture}[]

    \node [block, draw, text centered, rectangle, minimum height=1.2cm, fill=NextBlue!20, name=environment, rounded corners] {Environment};

    \node [block, text centered, rectangle, minimum height=1.2cm, right=1.2cm of environment, fill=NextBlue!20, name=adversary, rounded corners] {Adversary/\\Noise Process};

    \node [block, fill=NextBlue!60, above right=-1.55cm and 5.5cm of environment, dashed, rounded corners] (dqn) {DQN};
    \node (text7) [below =-.5cm of dqn, font=\scriptsize]  {$Q(\bm{s}_\text{adv}, \bm{a})$};
    \node (certified_defense) [above =0.2cm of  dqn]  {Robustness Analysis}; 
    \node [block, fill=NextBlue!60, above right=-.85cm and 2.8cm of dqn, rounded corners] (text4) {Robust\\Action\\Selection};
    \begin{scope}[on background layer]
    \node [container,fit=(dqn) (certified_defense) (text4),inner sep=0.3cm, rounded corners] (container) {};
    \end{scope}
    \begin{scope}[on background layer]
    \node[fit= (certified_defense) (dqn),draw,fill=NextBlue!60, inner sep=0.1cm, rounded corners] (certified_defense_box)   {};
    \end{scope}
    \node (text6) [above right=0.6cm and 1.0cm of dqn]  {\textbf{Agent}};

    \draw [-latex,thick] (environment) -- (adversary) node[align=center,draw=none,fill=none,pos=0.4,below]{True\\State,\\$\bm{s}_{0}$};

    \node (eps_adv) [above=1.0cm of adversary,align=center]{Perturbation Parameters,\\$\{\bm{\epsilon}_\text{adv}, p_\text{adv}\}$ or $\{\bm{\sigma}\}$};
    \draw [-latex,thick] (eps_adv) -- (adversary.north) node[align=center,draw=none,fill=none]{};

    \node (eps_rob) [align=center] at (eps_adv -| certified_defense_box){Robustness Parameters,\\$\{\bm{\epsilon}_\text{rob}, p_\text{rob}\}$};
    \draw [-latex,thick] (eps_rob) -- (certified_defense_box.north) node[align=center,draw=none,fill=none]{};

    \draw [-latex,thick] (adversary) -- (certified_defense_box) node[align=center,draw=none,fill=none,pos=0.4,below]{Obs.,\\$\bm{s}_\text{adv}$};

    \node (certificate) [right=2.3cm of text4]{};
    \draw [-latex,thick] (text4) -- (certificate) node[draw=none,fill=none,pos=0.4,below]{};
    \node (text_certificate) [below left=0.0cm and 0.2cm of certificate]  {Certificate};

    \draw [-latex,thick] (certified_defense_box) -- (text4) 
    node[align=center,draw=none,fill=none,midway,below]{Guaranteed\\Lower Bnds,\\$Q_l(\bm{s}_\text{adv},\bm{a})$};
    \draw [-latex,thick] (text4.south) --  ($ (text4.south) - (0cm,1.5cm) $) -- node[align=center,draw=none,fill=none,midway,above] {Certifiably Robust Action, $a_\text{CARRL}$} ($ (environment.south) - (0cm,1.5cm) $) -- ($ (environment.south) - (0,0) $); %
  \end{tikzpicture}
\caption[System Architecture for Certified Adversarial Robustness for Deep Reinforcement Learning]{
System Architecture.
During online execution, an agent observes a state, $\bm{s}_\text{adv}$, corrupted by an adversary or noise process (constrained to $\bm{s}_\text{adv}\in\mathcal{B}_{p_\text{adv}}(\bm{s}_0, \bm{\epsilon}_\text{adv})$.
A trained Deep \acrshort*{rl} algorithm, e.g., \acrshort*{dqn}~\citep{Mnih_2015}, predicts the state-action values, $Q$.
The robustness analysis module computes a lower bound of the network's predicted state-action values of each discrete action: $Q_l$, w.r.t. a robustness threshold $\bm{\epsilon}_\text{rob}$ and $\ell_{p_\text{rob}}$-norm in the input space.
The agent takes the action, $a_\text{CARRL}$, that maximizes the lower bound, i.e., is best under a worst-case deviation in the input space, and returns a \textit{certificate} bounding $a_\text{CARRL}$'s sub-optimality.
}
  \label{fig:sys_arch} 
\end{figure*}
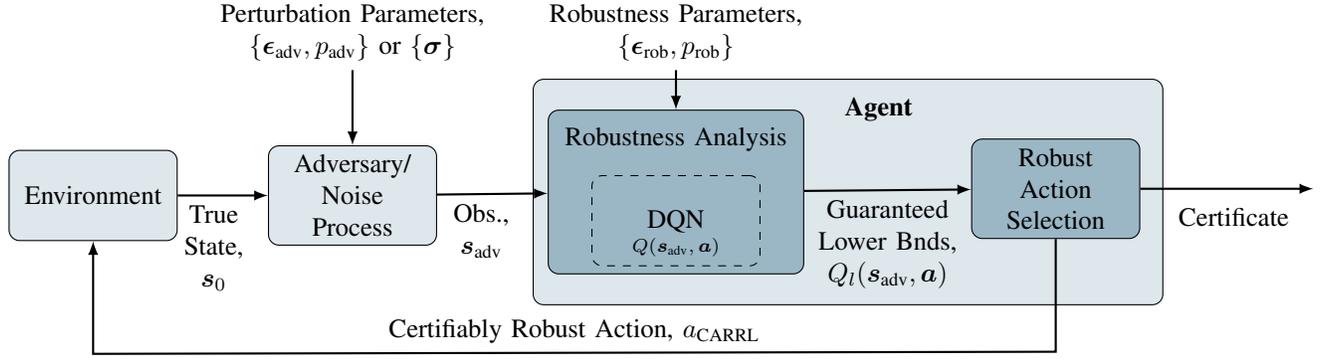

In an offline training phase, an agent uses a deep \acrshort*{rl} algorithm, here \acrshort*{dqn}~\citep{Mnih_2015}, to train a \acrshort*{dnn} that maps non-corrupted state observations, $\bm{s}_0$, to state-action values, $Q(\bm{s}_0,a)$.
Action selection during training uses the nominal cost function, $a^*_\text{nom} = \argmax_{a_j} Q(\bm{s}_0, a_j)$.

\Cref{fig:sys_arch} depicts the system architecture of a standard model-free \acrshort*{rl} framework with the added-on robustness module.
During online execution, the agent only receives corrupted state observations from the environment.
The robustness analysis node uses the \acrshort*{dnn} architecture, \acrshort*{dnn} weights, $W$, and robustness hyperparameters, $\bm{\epsilon}_\text{rob}, p_\text{rob}$, to compute lower bounds on possible Q-values for robust action selection.

\subsection{Optimal cost function under worst-case perturbation} \label{sec:approach:optimal_cost_function}
We assume that the training process causes the network to converge to the optimal value function, $Q^*(\bm{s}_0,a)$ and focus on the challenge of handling perturbed observations during execution.
Thus, we consider robustness to an adversary that perturbs the true state, $\bm{s}_0$, within a small perturbation, $\bm{\epsilon}_\text{adv}$, into the worst-possible state observation, $\bm{s}_\text{adv}$.
The adversary assumes that the \acrshort*{rl} agent follows a nominal policy (as in, e.g., \acrshort*{dqn}) of selecting the action with highest Q-value at the current observation.
A worst possible state observation, $\bm{s}_\text{adv}$, is therefore any one which causes the \acrshort*{rl} agent to take the action with lowest Q-value in the true state, $\bm{s}_0$:
\begin{align}
\bm{s}_\text{adv} \in \{\bm{s}:\; &\bm{s} \in \mathcal{B}_{p_\text{adv}}(\bm{s}_0, \bm{\epsilon}_\text{adv})\, \text{and}\,\nonumber \\&
\argmax_{a_j} Q(\bm{s}, a_j) = \argmin_{a_j} Q(\bm{s}_0, a_j)\}.
\label{eq:s_adv}
\end{align}
This set could be computationally intensive to compute and/or empty -- an approximation is described in~\cref{sec:approach:adversaries}.

After the agent receives the state observation picked by the adversary, the agent selects an action.
Instead of trusting the observation (and thus choosing the worst action for the true state), the agent could leverage the fact that the true state, $\bm{s}_0$, must be somewhere inside an $\bm{\epsilon}_\text{adv}$-Ball around $\bm{s}_\text{adv}$ (i.e., ${\bm{s}_0 \in \mathcal{B}_{p_\text{adv}}(\bm{s}_\text{adv}, \bm{\epsilon}_\text{adv}))}$.

However, in this work, the agent assumes ${\bm{s}_0 \in \mathcal{B}_{p_\text{rob}}(\bm{s}_\text{adv}, \bm{\epsilon}_\text{rob}))}$, where we make a distinction between the adversary and robust agent's parameters.
This distinction introduces hyperparameters $\bm{\epsilon}_\text{rob},\,p_\text{rob}$ that provide further flexibility in the defense algorithm's conservatism, that do not necessarily have to match what the adversary applies.
Nonetheless, for the sake of providing guarantees, the rest of~\cref{sec:approach} assumes ${\bm{\epsilon}_\text{rob}=\bm{\epsilon}_\text{adv}}$ and ${p_\text{rob}=p_\text{adv}}$ to ensure ${\mathcal{B}_{p_\text{rob}}(\bm{s}_\text{adv}, \bm{\epsilon}_\text{rob}))=\mathcal{B}_{p_\text{adv}}(\bm{s}_\text{adv}, \bm{\epsilon}_\text{adv}))}$.
Empirical effects of tuning ${\bm{\epsilon}_\text{rob}}$ to other values are explored in~\cref{sec:results}.

The agent evaluates each action by calculating the worst-case Q-value under all possible true states.
\begin{definition}
In accordance with the robust decision making problem, the robust-optimal action, $a^*$, is defined here as one with the highest Q-value under the worst-case perturbation,
\begin{equation}
 a^* = \argmax_{a_j}\underbrace{\min_{\bm{s}\in \mathcal{B}_{p_\text{rob}}(\bm{s}_\text{adv}, \bm{\epsilon}_\text{rob})} Q(\bm{s},a_j)}_{Q_L(\bm{s}_\text{adv}, a_j)}.\label{eq:robust_optimal_action_rule}
\end{equation}
\end{definition}
As described in~\cref{sec:related_work}, computing $Q_L(\bm{s}_\text{adv},a_j)$ exactly is too computationally intensive for real-time decision-making.

Thus, this work proposes the algorithm \textbf{C}ertified \textbf{A}dversarial \textbf{R}obustness for Deep \textbf{RL} (\textbf{CARRL}).
\begin{definition}
In \acrshort*{carrl}, the action, $a_\text{CARRL}$, is selected by approximating $Q_L(\bm{s}_\text{adv},a_j)$ with $Q_l(\bm{s}_\text{adv}, a_j)$, its guaranteed lower bound across all possible states $\bm{s}\in\mathcal{B}_{p_\text{rob}}(\bm{s}_\text{adv},\bm{\epsilon}_\text{rob})$, so that:
\begin{equation}
 a_\text{CARRL} = \argmax_{a_j} Q_l(\bm{s}_\text{adv}, a_j),
\label{eq:opt_cost_fn} 
\end{equation}
where \cref{eq:qlj_norm} below defines $Q_l(\bm{s}_\text{adv}, a_j)$ in closed form.
\end{definition}
Conditions for optimality (${a^*=a_\text{CARRL}}$) are described in~\cref{sec:approach:guarantees}.

\subsection{Robustness certification with vector-$\mathbf{\epsilon}$-ball perturbations}\label{sec:approach:variable_e}
To solve~\cref{eq:opt_cost_fn} when $Q(\bm{s},a_j)$ is represented by a \acrshort*{dnn}, we adapt the formulation from~\citep{Weng_2018}.
Most works in adversarial examples, including~\citep{Weng_2018}, focus on perturbations on image inputs, in which all channels have the same scale (e.g., grayscale images with pixel intensities in $[0,255]$).
More generally, however, input channels could be on different scales (e.g., joint torques, velocities, positions).
Existing robustness analysis methods require choosing a scalar $\epsilon_\text{rob}$ that bounds the uncertainty across all input channels; in general, this could lead to unnecessarily conservative behavior, as some network inputs might be known with zero uncertainty, or differences in units could make uncertainties across channels incomparable.
Hence, this work computes bounds on the network output under perturbation bounds specific to each network input channel, as described in a vector $\bm{\epsilon}_\text{rob}$ with the same dimension as $\bm{s}$.

To do so, we minimize $\bar{Q}_l(\bm{s},a_j)$ across \textit{all} states in ${\mathcal{B}_{p_\text{rob}}(\bm{s}_\text{adv}, \bm{\epsilon}_\text{rob})}$, where $\bar{Q}_l(\bm{s},a_j)$ was defined in \cref{eq:q_certified_lower_bound_closed_form} as the lower bound on the Q-value for a \textit{particular} state ${\bm{s}\in\mathcal{B}_{p_\text{rob}}(\bm{s}_\text{adv}, \bm{\epsilon}_\text{rob})}$.
This derivation uses a vector $\bm{\epsilon}_\text{rob}$ (instead of scalar $\epsilon_\text{rob}$ as in~\cite{Weng_2018}):
\allowdisplaybreaks
\begin{align}
\begin{split}
  Q_l(\bm{s}_\text{adv}, a_j) &=\min_{\bm{s}\in \mathcal{B}_p(\bm{s}_\text{adv},\bm{\epsilon}_\text{rob})} \left(\vphantom{\sum_{a}^{b}} \bar{Q}_l(\bm s, a_j) \right) \label{eq:rec}
\end{split}\\
\begin{split}
 &= \min_{\bm{s}\in \mathcal{B}_p(\bm{s}_\text{adv},\bm{\epsilon}_\text{rob})} \left(\vphantom{\sum_{a}^{b}} A_{j,:}^{(0)}\bm{s} + \right. \\*& 
 \quad\quad\quad\quad\left. \underbrace{b_j^{(m)} + \sum_{k=1}^{m-1}{A_{j,:}^{(k)}(\bm{b}^{(k)}-H_{:,j}^{(k)})}}_{=: \Gamma}\right)
 \label{eq:qlj_full}
 \end{split}
\raisetag{3\baselineskip}\\
\begin{split}
& = \min_{\bm{s}\in \mathcal{B}_p(\bm{s}_\text{adv},\bm{\epsilon}_\text{rob})}\left(A_{j,:}^{(0)}\bm{s}\right) + \Gamma \label{eq:qlj_s}
\end{split}\\
\begin{split}
& = \min_{\bm{y}\in \mathcal{B}_p(\bm{0}, \bm{1})}\left(A_{j,:}^{(0)} (\bm{y}\odot\bm{\epsilon}_\text{rob})\right) + A_{j,:}^{(0)}\bm{s}_\text{adv} + \Gamma\label{eq:qlj_y}
\end{split}\\
\begin{split}
& = \min_{\bm{y}\in \mathcal{B}_p(\bm{0}, \bm{1})}\left((\bm{\epsilon}_\text{rob}\odot A_{j,:}^{(0)}) \bm{y}\right) + A_{j,:}^{(0)}\bm{s}_\text{adv} + \Gamma\label{eq:qlj_shuffle}
\end{split}\\
\begin{split}
&= -\lvert\lvert\bm{\epsilon}_\text{rob}\odot A_{j,:}^{(0)}\rvert\rvert_q + A_{j,:}^{(0)}\bm{s}_\text{adv} + \Gamma\label{eq:qlj_norm},
\end{split}
\raisetag{2.5\baselineskip}
\end{align}
with $\odot$ denoting element-wise multiplication.
From~\cref{eq:rec} to~\cref{eq:qlj_full}, we substitute in~\cref{eq:q_certified_lower_bound_closed_form}.
From~\cref{eq:qlj_full} to~\cref{eq:qlj_s}, we introduce the placeholder variable $\Gamma$ that does not depend on $\bm s$.
From~\cref{eq:qlj_s} to~\cref{eq:qlj_y}, we substitute ${\bm{s} := \bm{y} \odot \bm{\epsilon}_\text{rob} + \bm{s}_\text{adv}}$, to shift and re-scale the observation to within the unit ball around zero, $\bm{y}\in \mathcal{B}_p(\bm{0},\bm{1})$.
The maximization in~\cref{eq:qlj_shuffle} is equivalent to a $\ell_q$-norm in~\cref{eq:qlj_norm} by the definition of the dual norm $\lvert\lvert \bm{z} \rvert\rvert_q = \{\text{sup}\ \bm{z}^T \bm{y}:\lvert\lvert \bm{y} \rvert\rvert_p \leq 1\}$ and the fact that the $\ell_q$ norm is the dual of the $\ell_p$ norm for $p,q \in [1,\infty)$ (with ${1/p + 1/q = 1}$).
\Cref{eq:qlj_norm} is inserted into~\cref{eq:opt_cost_fn} to calculate the \acrshort*{carrl} action in closed form.

Recall from~\cref{sec:related_work} that the bound $Q_l$ is the greedy (one linear upper and lower bound per activation) solution of an \acrshort*{lp} that describes a \acrshort*{dnn} with \acrshort*{relu} activations.
In this work, we refer to the full (non-greedy) solution to the primal convex relaxed \acrshort*{lp} as $Q_{LP}$ (cf. Problem $\mathcal{C}$, \textbf{LP-All} in~\citep{salman2019convex}).

To summarize, the relationship between each of the Q-value terms is:
\begin{align}
    Q(\bm{s}_\text{adv}, a_j) &\geq Q_L(\bm{s}_\text{adv}, a_j) \geq Q_{LP}(\bm{s}_\text{adv}, a_j) \geq Q_l(\bm{s}_\text{adv}, a_j) \label{eq:Q_geq_QL_geq_QLP_geq_Ql}\\
    Q(\bm{s}_\text{adv}, a_j) &\geq \bar{Q}_l(\bm{s}_\text{adv}, a_j) \geq Q_l(\bm{s}_\text{adv}, a_j) \label{eq:Q_geq_Qlbar_geq_Ql}.
\end{align}

\subsection{Guarantees on Action Selection}\label{sec:approach:guarantees}
\subsubsection{Avoiding a Bad Action}
Unlike the nominal DQN rule, which could be tricked to take an arbitrarily bad action, the robust-optimal decision-rule in~\cref{eq:robust_optimal_action_rule} can avoid bad actions provided there is a better alternative in the $\bm{\epsilon}_\text{rob}$-Ball.

\textbf{Claim 1:} If for some $q'$, $\exists \bm{s}' \in \mathcal{B}_{p_\text{rob}}(\bm{s}_\text{adv}, \bm{\epsilon}_\text{rob}), a'$ s.t. $Q(\bm{s}',a')\leq q'$ and $\exists a''$ s.t. $\forall \bm{s}\in\mathcal{B}_{p_\text{rob}}(\bm{s}_\text{adv}, \bm{\epsilon}_\text{rob})\ Q(\bm{s},a'')>q'$, then $a^* \neq a'$.

In other words, if action $a'$ is sufficiently bad for some nearby state, $\bm{s}'$, and at least one other action $a''$ is better for all nearby states, the robust-optimal decision rule will not select the bad action $a'$ (but DQN might).
This is because $Q_L(\bm{s}_\text{adv},a')\leq q'$, $Q_L(\bm{s}_\text{adv}, a'')>q' \Rightarrow \argmax_{a_j}Q_L(\bm{s}_\text{adv}, a_j) \neq a'$.

\subsubsection{Matching the Robust-Optimal Action}
While Claim 1 refers to the robust-optimal action, the action returned by \acrshort*{carrl} is the same as the robust-optimal action (returned by a system that can compute the exact lower bounds) under certain conditions,
\begin{equation}
  \underbrace{\argmax_{a_j} Q_l(\bm{s}_\text{adv}, a_j)}_{a_\text{CARRL}} \stackrel{?}{=} \underbrace{\argmax_{a_j} Q_L(\bm{s}_\text{adv}, a_j)}_{a^*}.   
\end{equation}

\textbf{Claim 2:} \acrshort*{carrl} selects the robust-optimal action if the robustness analysis process satisfies $Q_l = g(Q_L)$, where $g$ is a strictly monotonic function,
where $Q_l, Q_L$ are written without their arguments, $\bm{s}_\text{adv}, a_j$.
A special case of this conditions is when the analysis returns a tight bound, i.e., $Q_l(\bm{s}_\text{adv}, a_j) = Q_L(\bm{s}_\text{adv}, a_j)$.
Using the Fast-Lin-based approach in this work, for a particular observation, confirmation that all of the ReLUs are ``active'' or ``inactive'' in \cref{eq:sigma_lk_uk} would provide a tightness guarantee.

In cases whereClaim 2 is not fulfilled, but Claim 1 is fulfilled, \acrshort*{carrl} is not guaranteed to select the robust-optimal action, but will still reason about all possible outcomes, and empirically selects a better action than a nominal policy across many settings explored in~\cref{sec:results}.

Note that when $\bm{\epsilon}_\text{rob}=\bm{0}$, no robustness is applied, so both the \acrshort*{carrl} and robust-optimal decisions reduce to the DQN decision, since $Q_l(\bm{s}_\text{adv}, a_j) = Q_L(\bm{s}_\text{adv}, a_j) = Q(\bm{s}_\text{adv}, a_j)$.

\textbf{Claim 3:} \acrshort*{carrl} provides a \textit{certificate} of the chosen action's sub-optimality,
\begin{align}
  0 &\leq Q(\mathbf{s}_0, \, a^{**}) - Q(\mathbf{s}_0, \, a_\text{CARRL}) \nonumber\\
  &\leq \underbrace{\left(\max_{a_j} Q_u(\mathbf{s}_\text{adv}, \, a_j)\right) - Q_l(\mathbf{s}_\text{adv}, \, a_\text{CARRL})}_{\text{Sub-optimality Certificate}},
  \label{eqn:carrl:suboptimality_certificate}
\end{align}
where $a^{**}=\max_{a_j} Q(\mathbf{s}_0,\,a_j)$, i.e., the Q-maximizing action for the true (unknown) state, and $Q_u$ is an upper bound on $Q$ computed analogously to $Q_l$ in~\cref{eq:qlj_norm}.

\begin{proof}
The definition of $a^{**}$ implies $Q(\mathbf{s}_0, \, a^{**}) \geq Q(\mathbf{s}_0, \, a_j)$ for any $a_j\in\mathcal{A}$, including $a_j=a_\text{CARRL}$.
This proves the lower bound.

For any $a_j\in\mathcal{A}$ including $a_j=a_\text{CARRL}$, the upper bound analog of~\cref{eq:Q_geq_QL_geq_QLP_geq_Ql} gives $Q_u(\mathbf{s}_\text{adv}, \, a_j) \geq {Q_U(\mathbf{s}_\text{adv}, \, a_j) = \max_{\mathbf{s}\in\mathcal{B}_{p_\text{rob}}(\bm{s}_\text{adv}, \bm{\epsilon}_\text{rob})} Q(\mathbf{s}, \, a_j) \geq Q(\mathbf{s}_0, \, a_j)}$.
Using the definition of $a^{**}$ thus gives ${\max_{a_j} Q_u(\mathbf{s}_\text{adv}, \, a_j) \geq \max_{a_j} Q(\mathbf{s}_0, \, a_j) = Q(\mathbf{s}_0, \, a^{**})}$.
Using~\cref{eq:Q_geq_QL_geq_QLP_geq_Ql,eq:robust_optimal_action_rule}, ${Q_l(\mathbf{s}_\text{adv}, \, a_j) \leq Q_L(\mathbf{s}_\text{adv}, \, a_j) \leq Q_l(\mathbf{s}_0, \, a_j)}$, because $\mathbf{s}_0\in\mathcal{B}_{p_\text{rob}}(\bm{s}_\text{adv}, \bm{\epsilon}_\text{rob})$.
This proves the upper bound.
\end{proof}

The benefit of \cref{eqn:carrl:suboptimality_certificate} is that it bounds the difference in the Q-value of CARRL's action under input uncertainty and the action that would have been taken given knowledge of the true state.
Thus, the bound in~\cref{eqn:carrl:suboptimality_certificate} captures the uncertainty in the true state (considering the whole $\mathbf{\epsilon}$-ball through $Q_u, Q_l$), and the unknown action $a^{**}$ that is optimal for the true (unknown) state.

This certificate of sub-optimality is what makes CARRL \textit{certifiably robust}, with further discussion in~\cref{sec:cert_vs_verif}.

\subsection{Probabilistic Robustness}
The discussion so far considered cases where the state perturbation is known to be bounded (as in, e.g., many adversarial observation perturbation definitions~\citep{Goodfellow_2015}, stochastic processes with finite support.
However, in many other cases, the observation uncertainty is best modeled by a distribution with infinite support (e.g., Gaussian).
To be fully robust to this class of uncertainty (including very low probability events) \acrshort*{carrl} requires setting $\epsilon_{rob,i}=\infty$ for the unbounded state elements, $\bm{s}_i$.

For instance, for a Gaussian sensor model with known standard deviation of measurement error, $\bm{\sigma}_{sensor}$, one could set ${\bm{\epsilon}_\text{rob}=2\bm{\sigma}_{sensor}}$ to yield actions that account for the worst-case outcome with 95\% confidence.
In robotics applications, for example, $\bm{\sigma}_{sensor}$ is routinely provided in a sensor datasheet.
Implementing this type of probabilistic robustness only requires a sensor model for the observation vector and a desired confidence bound to compute the corresponding $\bm{\epsilon}_\text{rob}$.

\subsection{Design of a Policy with Reduced Sensitivity}\label{sec:approach:reduced_sensitivity_policy}
While~\cref{eq:robust_optimal_action_rule} provides a robust policy by selecting an action with best worst-case performance, an alternative robustness paradigm could prefer actions with low sensitivity to changes in the input while potentially sacrificing performance.
Using the same machinery as before to compute guaranteed output bounds per action, we add a cost term to penalize sensitivity.
This interpretation gives a reduced sensitivity (RS) action-selection rule,
\begin{align}
a^{*}_\text{RS} &= \argmax_{a_j} \left[\underbrace{Q_L(\bm{s}_\text{adv},a_j)}_\text{Worst-Case Performance} - \right. \label{eq:nonsens_performance_action_rule} \\ \nonumber
& \quad\quad \left. \lambda_\text{sens}\cdot \underbrace{\left( Q_U(\bm{s}_\text{adv}, a_j) - Q_L(\bm{s}_\text{adv}, a_j) \right)}_\text{Sensitivity}\right],
\end{align}%
where $Q_U$ is defined analogously to $Q_L$ in~\cref{eq:robust_optimal_action_rule} and $\lambda_\text{sens}$ is a hyperparameter to trade-off nominal performance and sensitivity.
In practice, guaranteed bounds $Q_u, \, Q_l$ can be used instead of $Q_U, \, Q_L$ in~\cref{eq:nonsens_performance_action_rule} for a coarser estimate of action sensitivity.

The benefit of the formulation in~\cref{eq:nonsens_performance_action_rule} is that it penalizes actions with a wide range of possible Q-values for the input set, while still considering worst-case performance.
That is, an action with a wide range of possible Q-values will be less desirable than in the previous robust action-selection rule in~\cref{eq:robust_optimal_action_rule}.
In the two extremes, $\lambda_\text{sens}=0 \implies a^{*}_\text{RS} = a^{*}$; whereas $\lambda_\text{sens}\to\infty \implies a^{*}_\text{RS}$ ignores performance and always chooses action that is least sensitive to input changes.

\subsection{Robust Policy-Based RL}\label{sec:approach:policy_based_rl}
There are two main classes of RL methods: value-function-based and policy-based.
The preceding sections introduce robustness methods for value-function-based RL algorithms, such as DQN~\citep{Mirman_2019}.
The other main class of RL methods instead learn a policy distribution directly, with or without simultaneously learning to estimate the value function (e.g., A3C~\citep{mnih2016asynchronous}, PPO~\citep{schulman2017proximal}, SAC~\citep{haarnoja2018soft}).
Here we show that the proposed robust optimization formulation can be adapted to policy-based RL with discrete actions.

For example, consider the Soft Actor-Critic (SAC)~\citep{haarnoja2018soft} algorithm (extended to discrete action spaces~\citep{christodoulou2019soft}).
After training, SAC returns a \acrshort*{dnn} policy.
We can use the trained policy logits as an indication of desirability for taking an action (last layer before softmax/normalization).
Let the logits be denoted as $\mu(\cdot\lvert\bm{s})\in\mathbb{R}^d$ and the policy distribution be $\pi(\cdot\lvert\bm{s})=\mathrm{softmax}(\mu(\cdot\lvert\bm{s}))\in\Delta^d$, where $\Delta^d$ denotes the standard $d$-simplex.

In accordance with the robust decision making problem, the robust-optimal action, $a^*_\pi$, is defined here as one with the highest policy logit under the worst-case perturbation,
\begin{equation}
a^*_\pi = \argmax_{a_j}\underbrace{\min_{\bm{s}\in \mathcal{B}_{p_\text{rob}}(\bm{s}_\text{adv}, \bm{\epsilon}_\text{rob})} \mu(a_j\lvert\bm{s})}_{\mu_L(a_j\lvert\bm{s}_\text{adv})}.\label{eq:robust_optimal_action_rule_policy}
\end{equation}
Our computationally tractable implementation selects $a_{\text{CARRL},\pi}$ by approximating $\mu_L(a_j\lvert\bm{s}_\text{adv})$ with $\mu_l(a_j\lvert\bm{s}_\text{adv})$, its guaranteed lower bound across all possible states $\bm{s}\in\mathcal{B}_{p_\text{rob}}(\bm{s}_\text{adv},\bm{\epsilon}_\text{rob})$, so that:
\begin{equation}
 a_{\text{CARRL},\pi} = \argmax_{a_j} \mu_l(a_j\lvert\bm{s}_\text{adv}),
\label{eq:opt_cost_fn_policy} 
\end{equation}
where $\mu_l(a_j\lvert\bm{s}_\text{adv})$ can be computed in closed form using~\cref{eq:qlj_norm}, where the only modification is the interpretation of the \acrshort*{dnn} output (Q-value vs. policy logits).

The robust, deterministic policy returned by this formulation applies to several standard RL methods, including~\citep{haarnoja2018soft,babaeizadeh2016ga3c}.
If a stochastic policy is preferred for other RL algorithms, one could sample from the normalized lower bounded logits, i.e., $a_{\text{CARRL},\pi} \sim \mathrm{softmax}(\mu_l(\cdot\lvert\bm{s}_\text{adv}))$.
Because of the similarity in formulations for robust policy-based and value-function-based RL, DQN is used as a representative example going forward.

\subsection{Adversaries}\label{sec:approach:adversaries}

To evaluate the learned policy's robustness to deviations of the input in an $\bm{\epsilon}_\text{adv}$-Ball, we pass the true state, $\bm{s}_0$, through an adversarial/noise process to compute $\bm{s}_\text{adv}$, as seen in~\cref{fig:sys_arch}.
Computing an adversarial state $\bm{s}_\text{adv}$ exactly, using \cref{eq:s_adv}, is computationally intensive.
Instead, we use a \acrfull*{fgst}~\citep{Kurakin_2017b} to approximate the adversary from~\cref{eq:s_adv} (other adversaries could include PGD~\citep{Madry_2018}, ZOO~\citep{chen2017zoo}).
FGST chooses a state $\hat{\bm{s}}_\text{adv}$ on the ${\ell_{\infty}\ \bm{\epsilon}_\text{adv}}$-Ball's perimeter to encourage the agent to take the nominally worst action, $\argmin_{a_j} Q(\bm{s}_0, a_j)$.
Specifically, $\hat{\bm{s}}_\text{adv}$ is picked along the direction of sign of the lowest cross-entropy loss, $\mathcal{L}$, between $\bm{y}_\text{adv}$, a one-hot encoding of the worst action at $\bm{s}_0$, and $\bm{y}_\text{nom}$, the softmax of all actions' Q-values at $\bm{s}_0$.
The loss is taken w.r.t. the proposed observation $\bm{s}$:
\begin{align}
\bm{y}_\text{adv} & = [\mathds{1} \{a_i = {\mathop{\mathrm{argmin}}\nolimits}_{a_j} Q(\bm{s}_0, a_j)\}] \in \mathbbm{U}^{d} \\
Q_\text{nom} & = [Q(\bm{s}_0, a_j)\forall j \in \{0,\ldots,d-1\}] \in\mathbb{R}^{d}\\
\bm{y}_\text{nom} & = \mathrm{softmax}(Q_\text{nom}) \in\Delta^{d}\\
\hat{\bm{s}}_\text{adv} &= \bm{s}_0-\bm{\epsilon}_\text{adv}\odot\mathrm{sign}(\nabla_{\bm{s}} \mathcal{L}(\bm{y}_\text{adv}, \bm{y}_\text{nom})),
\label{eq:s_adv_hat}
\end{align}
where $\mathbbm{U}^{d}$ denotes a one-hot vector of dimension $d$, and $\Delta^{d}$ denotes the standard $d$-simplex.

In addition to adversarial perturbations, \cref{sec:results} also considers uniform noise perturbations, where ${\bm{s}_\text{adv}\sim \mathtt{Unif}\left(\left[\bm{s}_0-\bm{\sigma}, \bm{s}_0+\bm{\sigma}\right]\right)}$.

\section{Experimental Results} \label{sec:results}
\begin{figure*}[t]
	\vspace{-0.15in}
	\centering
	\begin{minipage}[t]{0.19\linewidth}
	\centering\includegraphics [trim=0 0 0 0, clip, width=1.0\textwidth, angle = 0]{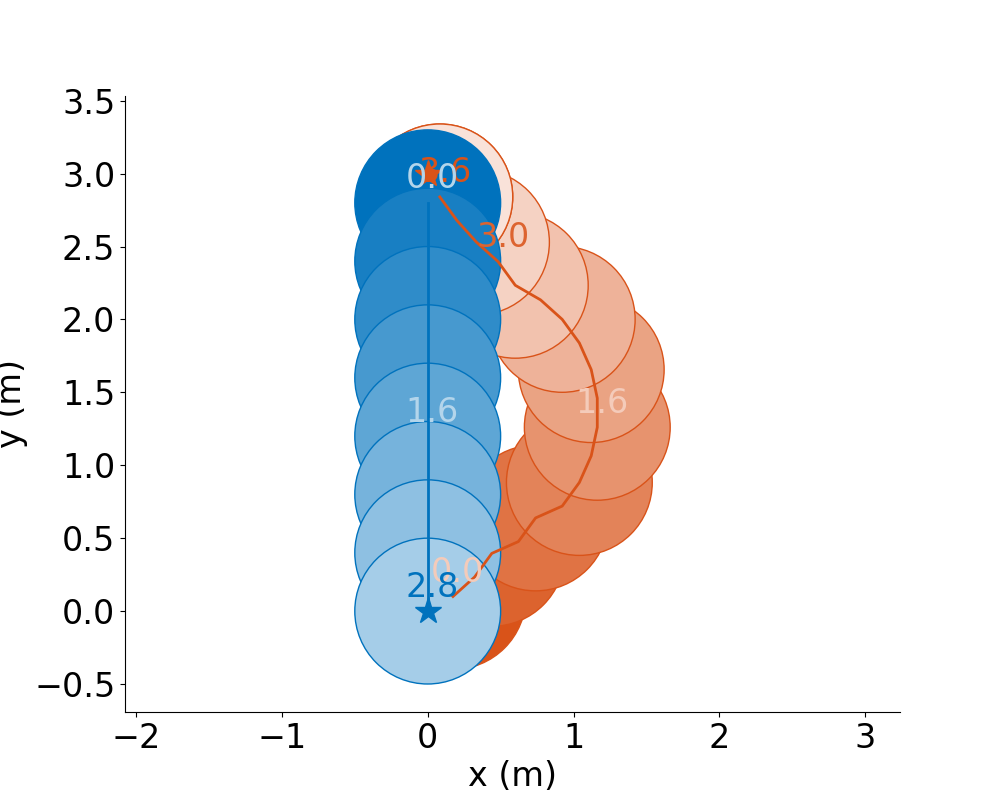}
	\subcaption{$\epsilon_\text{rob}=0.0$}\label{fig:collision_avoidance_traj_eps_0.0}
	\end{minipage}%
	\begin{minipage}[t]{0.19\linewidth}
	\centering\includegraphics [trim=0 0 0 0, clip, width=1.0\textwidth, angle = 0]{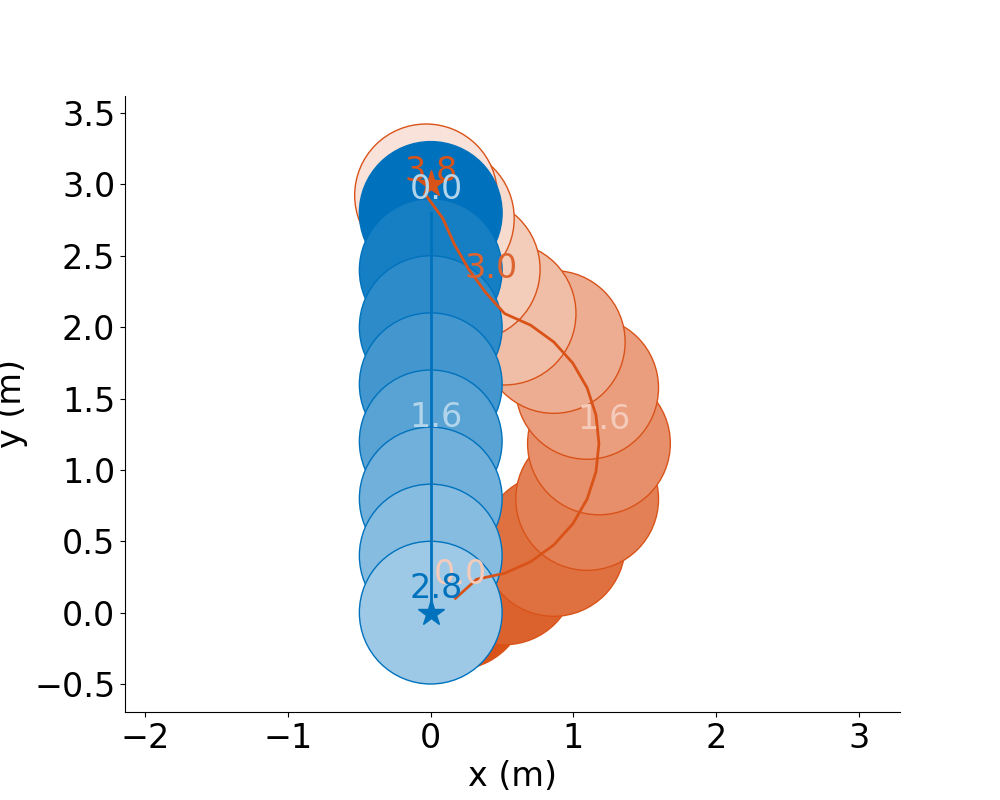}
	\subcaption{$\epsilon_\text{rob}=0.1$}\label{fig:collision_avoidance_traj_eps_0.1}
	\end{minipage}
	\begin{minipage}[t]{0.19\linewidth}
	\centering\includegraphics [trim=0 0 0 0, clip, width=1.0\textwidth, angle = 0]{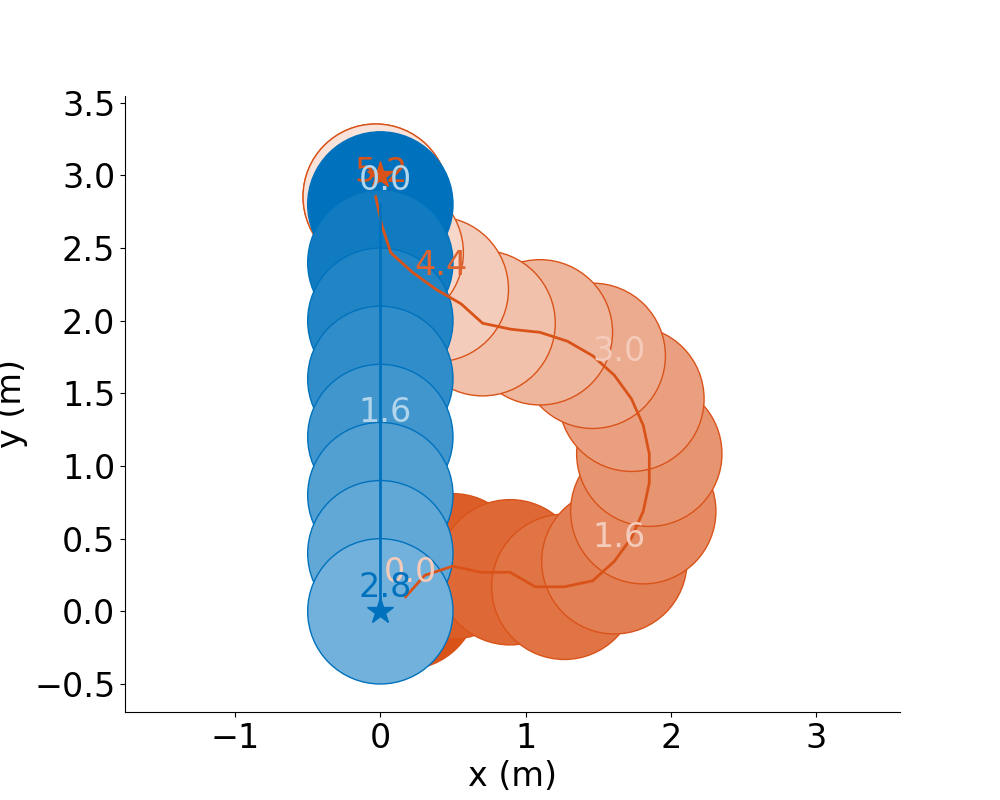}
	\subcaption{$\epsilon_\text{rob}=0.46$}\label{fig:collision_avoidance_traj_eps_0.46}
	\end{minipage}
	\begin{minipage}[t]{0.19\linewidth}
	\centering\includegraphics [trim=0 0 0 0, clip, width=1.0\textwidth, angle = 0]{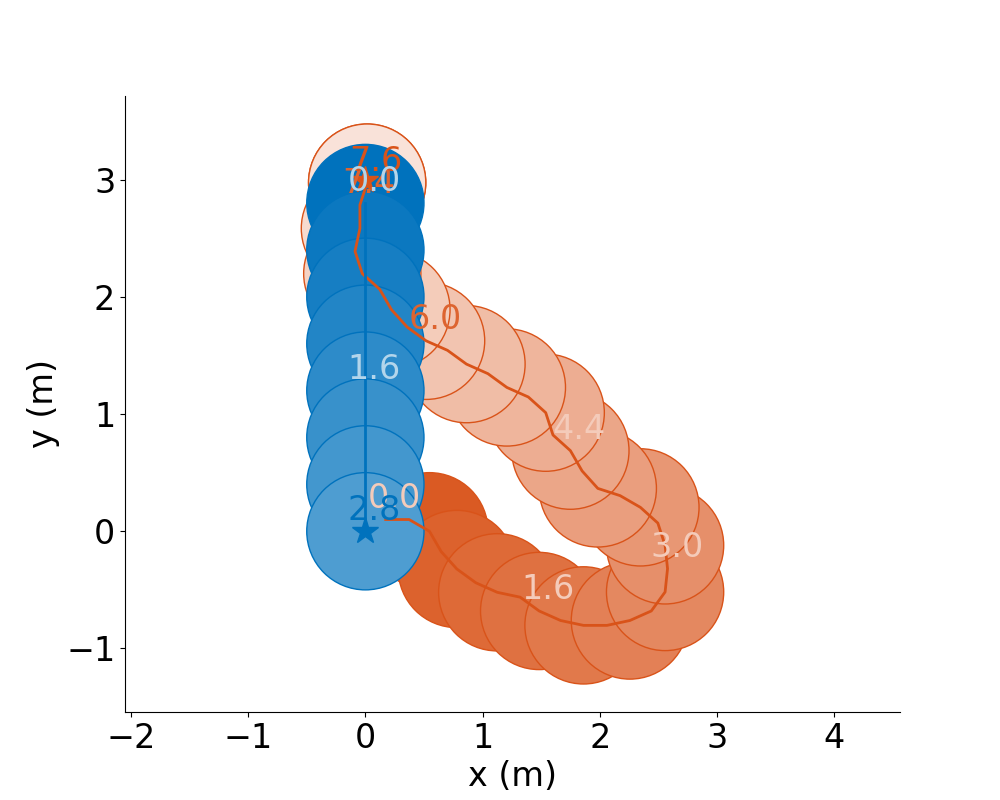}
	\subcaption{$\epsilon_\text{rob}=1.0$}\label{fig:collision_avoidance_traj_eps_1.0}
	\end{minipage}
	\begin{minipage}[t]{0.19\linewidth}
	\centering\includegraphics [trim=0 0 0 0, clip, width=1.0\textwidth, angle = 0]{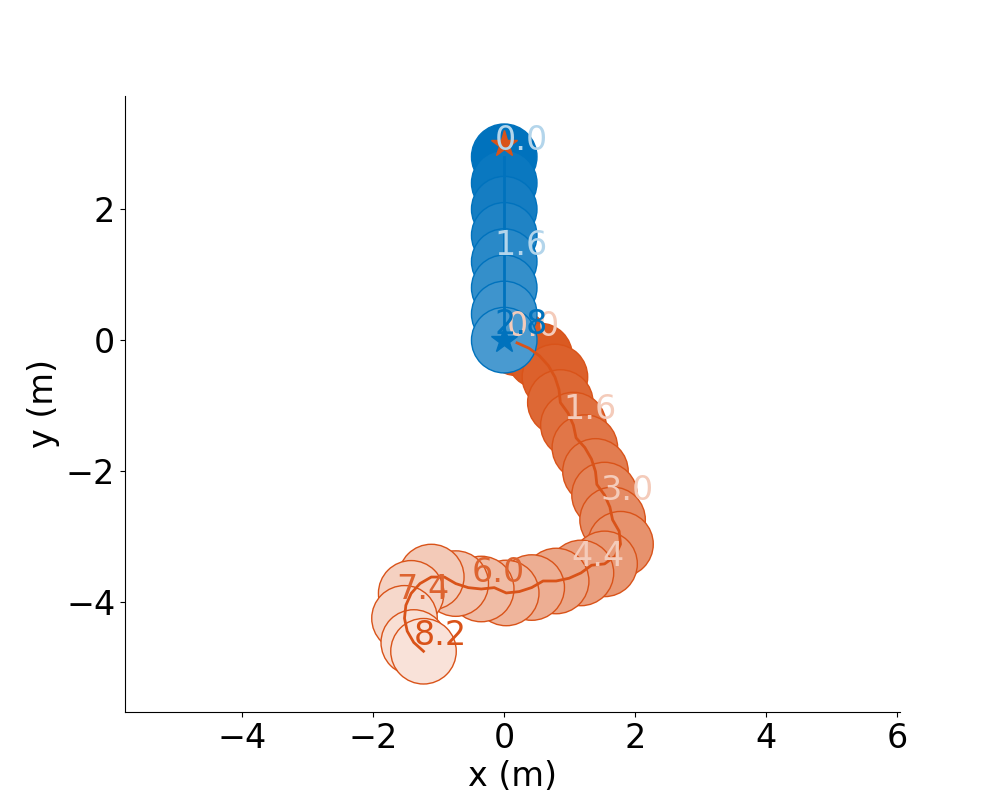}
	\subcaption{$\epsilon_\text{rob}=2.0$}\label{fig:collision_avoidance_traj_eps_2.0}
	\end{minipage}%
\caption[Increase of conservatism with increased $\epsilon_\text{rob}$ robustness]{Increase of conservatism with $\epsilon_\text{rob}$. An agent (orange) following the \acrshort*{carrl} policy avoids a dynamic, non-cooperative obstacle (blue) that is observed without noise. An increasing robustness parameter $\epsilon_\text{rob}$ (left to right) increases the agent's conservatism, i.e., the agent avoids the obstacle with a greater safety distance.}
\label{fig:collision_avoidance_trajs}
\end{figure*}

\begin{figure*}
	\centering
	\begin{minipage}[t]{0.33\linewidth}
	\centering\includegraphics [trim=0 0 0 0, clip, width=1.0\textwidth, angle = 0]{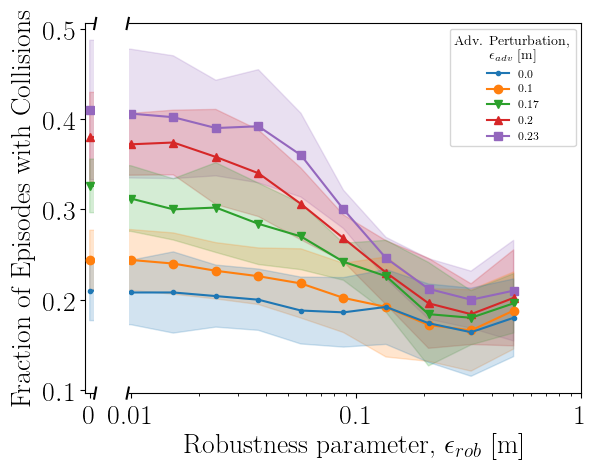}
	\vspace{-0.15in}
	\subcaption{}\label{fig:carrl_collision_avoidance_adv_colls}
	\end{minipage}%
	\begin{minipage}[t]{0.33\linewidth}
	\centering\includegraphics [trim=0 0 0 0, clip, width=1.0\textwidth, angle = 0]{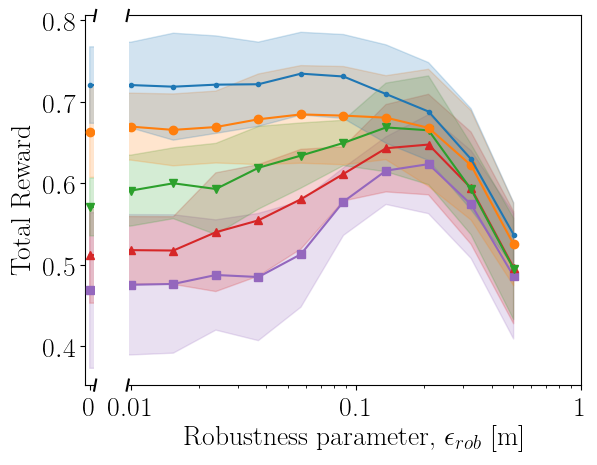}
	\vspace{-0.15in}
	\subcaption{}\label{fig:carrl_collision_avoidance_adv_rew}
	\end{minipage}
	\begin{minipage}[t]{0.33\linewidth}
	\centering\includegraphics [trim=0 0 0 0, clip, width=1.0\textwidth, angle = 0]{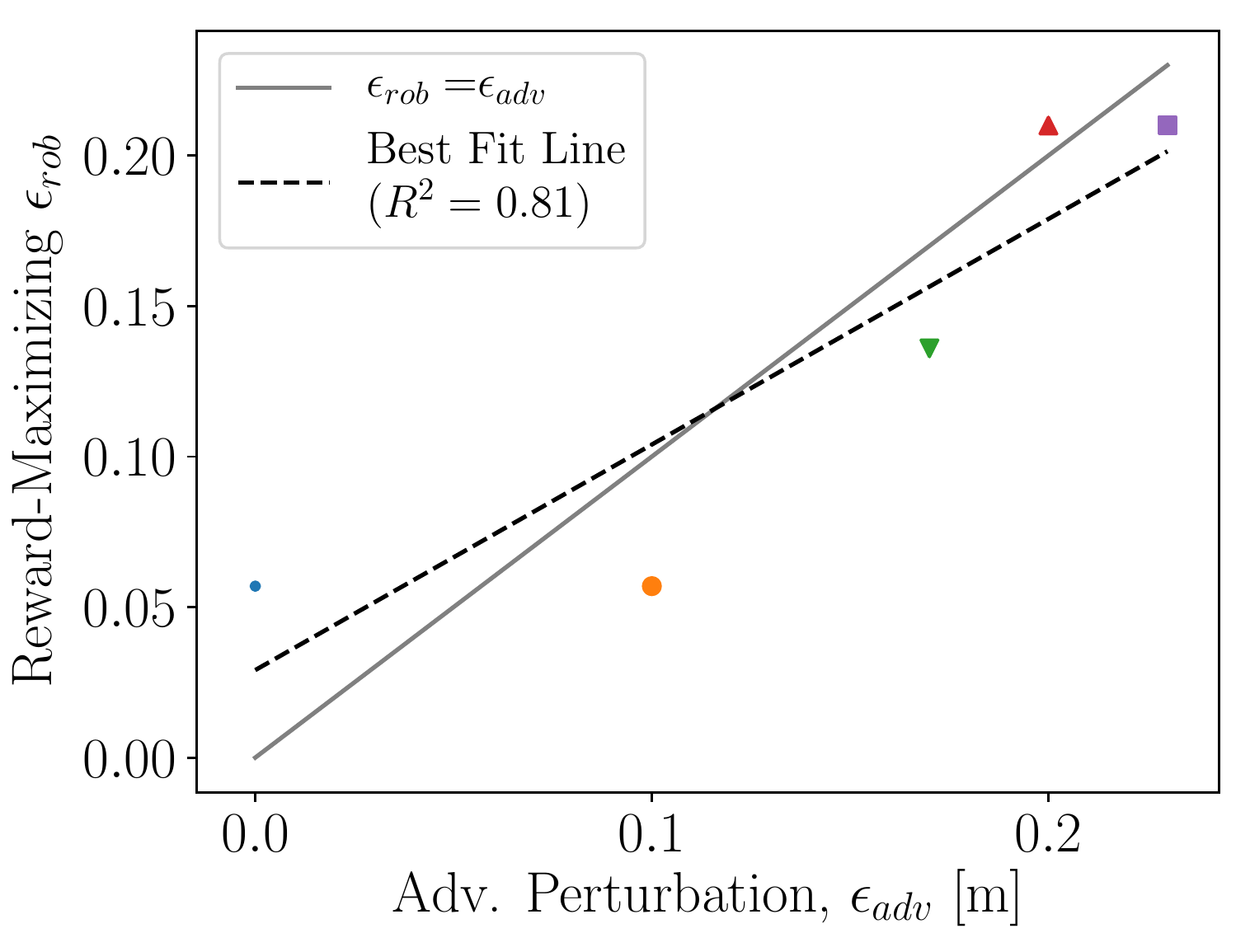}
	\subcaption{}\label{fig:carrl_collision_avoidance_adv_best_epsilon}
	\end{minipage}
	\\	
	\begin{minipage}[t]{0.33\linewidth}
	\centering\includegraphics [trim=0 0 0 0, clip, width=1.0\textwidth, angle = 0]{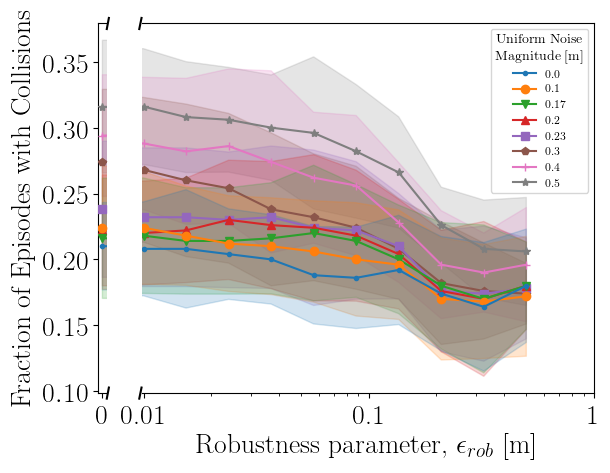}
	\vspace{-0.15in}
	\subcaption{}\label{fig:carrl_collision_avoidance_noise_colls}
	\end{minipage}%
	\begin{minipage}[t]{0.33\linewidth}
	\centering\includegraphics [trim=0 0 0 0, clip, width=1.0\textwidth, angle = 0]{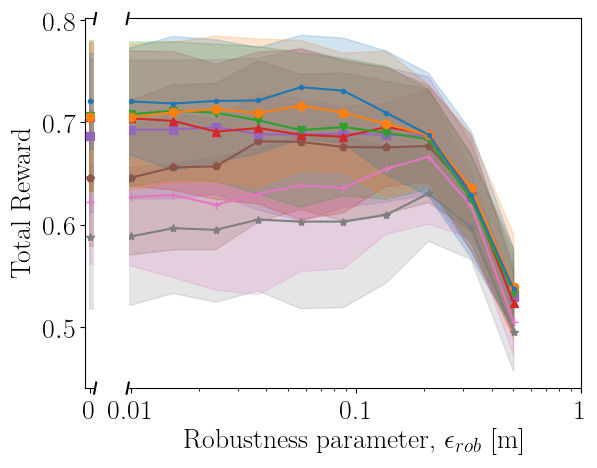}
	\vspace{-0.15in}
	\subcaption{}\label{fig:carrl_collision_avoidance_noise_rew}
	\end{minipage}
	\begin{minipage}[t]{0.33\linewidth}
	\centering\includegraphics [trim=0 0 0 0, clip, width=1.0\textwidth, angle = 0]{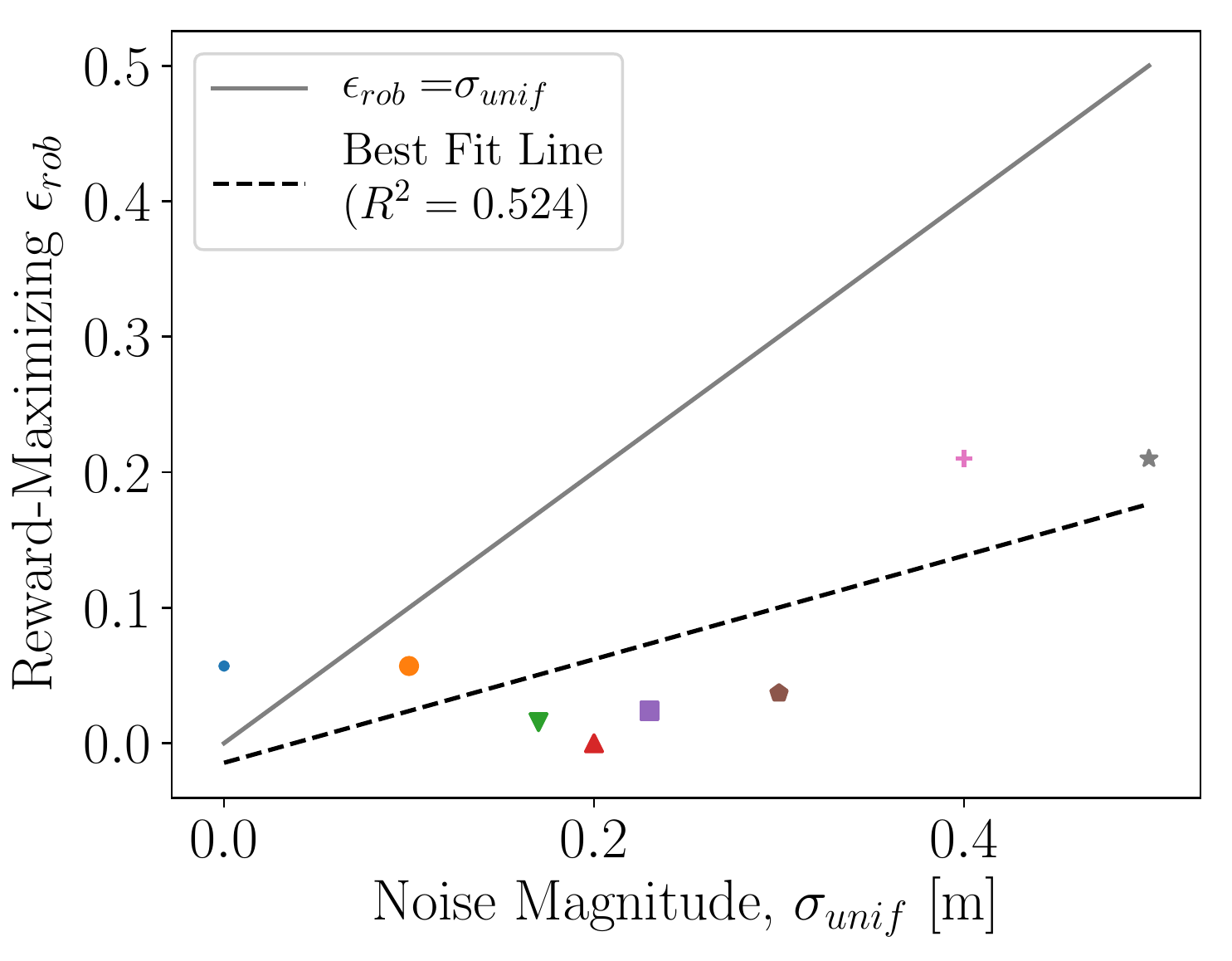}
	\subcaption{}\label{fig:carrl_collision_avoidance_noise_best_epsilon}
	\end{minipage}
	\vspace{-.05in}
\caption[Robustness against adversaries]{
Robustness against adversarial attacks (top row) and noise (bottom row).
Increasing the robustness parameter, $\epsilon_\text{rob}$, decreases the number of collisions in the presence of adversarial attacks (a), or noise (d) for various perturbation magnitudes, $\epsilon_\text{adv}, \sigma_{\text{unif}}$.
\acrshort*{carrl} also recovers substantial amounts of reward lost due to imperfect observations as $\epsilon_\text{rob}$ increases (b,e).
The choice of $\epsilon_\text{rob}$ to maximize reward in the presence of a particular adversary or noise process can be guided by the models in (c,f).
When $\epsilon_\text{rob}=0$, CARRL is equivalent to DQN.
}
\label{fig:carrl_collision_avoidance}
\end{figure*}

The key result is that while imperfect observations reduce the performance of a nominal deep \acrshort*{rl} algorithm, our proposed algorithm, \acrshort*{carrl}, recovers much of the performance by adding robustness during execution.
Robustness against perturbations from an adversary or noise process during execution is evaluated in two simulated domains: collision avoidance among dynamic, decision-making obstacles~\cite{gym_collision_avoidance} and cartpole~\cite{Brockman_2016}.
This section also describes the impact of the $\bm{\epsilon}_\text{rob}$ hyperparameter, the ability to handle behavioral adversaries, a comparison with another analysis technique, and provides intuition on the tightness of the lower bounds.

\subsection{Collision Avoidance Domain}\label{sec:results:collision_avoidance}

Among the many \acrshort*{rl} tasks, a particularly challenging safety-critical task is collision avoidance for a robotic vehicle among pedestrians.
Because learning a policy in the real world is dangerous and time consuming, this work uses a \textit{gym}-based~\cite{Brockman_2016} kinematic simulation environment~\cite{gym_collision_avoidance} for learning pedestrian avoidance policies.
In this work, the \acrshort*{rl} policy controls one of two agents with 11 discrete actions: change of heading angle evenly spaced between $[-\pi/6, +\pi/6]$ and constant velocity $v=1$ m/s.
The environment executes the selected action under unicycle kinematics, and controls the other agent from a diverse set of fixed policies (static, non-cooperative, \acrshort*{orca}~\citep{Berg_2009}, \acrshort*{ga3ccadrl}~\citep{Everett_2018}).
The sparse reward is $1$ for reaching the goal, $-0.25$ for colliding (and 0 otherwise, i.e., zero reward is received in cases where the agent does not reach its goal in a reasonable amount of time).
The observation vector includes the \acrshort*{carrl} agent's goal, each agent's radius, and the other agent's x-y position and velocity, with more detail in~\citep{Everett_2018}.
In this domain, robustness and perturbations are applied only on the measurement of the other agent's x-y position (i.e., ${\bm{\epsilon}_\text{rob}=\epsilon_\text{rob}\cdot\left[0\ \ldots\ 0\ 1\ 1\ 0\ \ldots0\right]}$) -- an example of \acrshort*{carrl}'s ability to handle uncertainties of varying scales.

A non-dueling \acrshort*{dqn} policy was trained with $2$, $64$-unit layers with the following hyperparameters: learning rate $2.05\texttt{e}{-4}$, $\epsilon$-greedy exploration ratio linearly decaying from $0.5$ to $0.05$, buffer size $152\texttt{e}3$, $4\texttt{e}5$ training steps, and target network update frequency, $10\texttt{e}3$.
The hyperparameters were found by running 100 iterations of Bayesian optimization with Gaussian Processes~\citep{Jasper_2012} on the maximization of the sparse training reward.

After training, \acrshort*{carrl} is added onto the trained \acrshort*{dqn} policy.
Intuition on the resulting \acrshort*{carrl} policy is demonstrated in~\cref{fig:collision_avoidance_trajs}.
The figure shows the trajectories of the \acrshort*{carrl} agent (orange) for increasing $\epsilon_\text{rob}$ values in a scenario with unperturbed observations.
With increasing $\epsilon_\text{rob}$ (toward right), the \acrshort*{carrl} agent accounts for increasingly large worst-case perturbations of the other agent's position.
Accordingly, the agent avoids the dynamic agent (blue) increasingly conservatively, i.e., selects actions that leave a larger safety distance.
When $\epsilon_\text{rob}{=}2.0$, the \acrshort*{carrl} agent is overly conservative -- it ``runs away'' and does not reach its goal, which is explained more in~\cref{sec:results:intuition_on_bounds}.

\Cref{fig:carrl_collision_avoidance} shows that the nominal \acrshort*{dqn} policy is not robust to the perturbation of inputs.
In particular, increasing the magnitude of adversarial, or noisy perturbation, $\epsilon_\text{adv}, \sigma_{\text{unif}}$, drastically
1) increases the average number of collisions (as seen in \cref{fig:carrl_collision_avoidance_adv_colls,fig:carrl_collision_avoidance_noise_colls}, respectively, at $\epsilon_\text{rob}=0$) and
2) decreases the average reward (as seen in \cref{fig:carrl_collision_avoidance_adv_rew,fig:carrl_collision_avoidance_noise_rew}).
The results in~\cref{fig:carrl_collision_avoidance} are evaluated in scenarios where the other agent is non-cooperative (i.e., travels straight to its goal position at constant velocity) and each datapoint represents the average of 100 trials across 5 random seeds that determine the agents' initial/goal positions and radii (shading represents $\pm1$ standard deviation of average value per seed).

Next, we demonstrate that \acrshort*{carrl} recovers performance.
Increasing the robustness parameter $\epsilon_\text{rob}$ decreases the number of collisions under varying magnitudes of noise, or adversarial attack, as seen in~\cref{fig:carrl_collision_avoidance_adv_colls,fig:carrl_collision_avoidance_noise_colls}.
Because collisions affect the reward function, the received reward also increases with an increasing robustness parameter $\epsilon_\text{rob}{<}{\sim}0.1$ under varying magnitudes of perturbations.
As expected, the effect of the proposed defense is highest under large perturbations, as seen in the slopes of the curves $\epsilon_\text{adv}{=}0.23$ (violet) and $\sigma_{\text{unif}}{=}0.5$ (gray).

Since the \acrshort*{carrl} agent selects actions more conservatively than a nominal \acrshort*{dqn} agent, it is able to successfully reach its goal instead of colliding like a nominal \acrshort*{dqn} agent does under many scenarios with noisy or adversarial perturbations.
However, the effect of overly conservative behavior seen in~\cref{fig:collision_avoidance_traj_eps_1.0,fig:collision_avoidance_traj_eps_2.0} appears in~\cref{fig:carrl_collision_avoidance} for $\epsilon_\text{rob}{>}\sim0.2$, as the reward drops significantly.
This excessive conservatism for large $\bm{\epsilon}_\text{rob}$ can be partially explained by the fact that highly conservative actions may move the agent's position observation into states that are far from what the network was trained on, which breaks \acrshort*{carrl}'s assumption of a perfectly learned Q-function.
Further discussion about the conservatism inherent in the lower bounds is discussed in~\cref{sec:results:intuition_on_bounds}.

\Cref{fig:carrl_collision_avoidance_adv_best_epsilon,fig:carrl_collision_avoidance_noise_best_epsilon} illustrate further intuition on choosing $\epsilon_\text{rob}$.
\Cref{fig:carrl_collision_avoidance_adv_best_epsilon} demonstrates a strong correlation between the attack magnitude $\epsilon_\text{adv}$ and the best (i.e., reward-maximizing) robustness hyperparameter $\epsilon_\text{rob}$ under that attack magnitude from~\cref{fig:carrl_collision_avoidance_adv_rew}.
In the case of uniform noise, the correlation between $\epsilon_\text{rob}$ and $\sigma_{\text{unif}}$ is weaker, because the \acrshort*{fgst} adversary chooses an input state on the perimeter of the $\bm{\epsilon}_\text{adv}$-Ball, whereas uniform noise samples lie inside the $\bm{\sigma}_{unif}$-Ball.

The flexibility in setting $\epsilon_\text{rob}$ enables \acrshort*{carrl} to capture uncertainties of various magnitudes in the input space, e.g., $\epsilon_\text{rob}$ could be adapted on-line to account for a perturbation magnitude that is unknown a priori, or to handle time-varying sensor noise.

\subsection{Cartpole Domain}\label{sec:results:cartpole}

\begin{figure*}
	\centering
	\begin{minipage}{0.45\linewidth}
	\centering\includegraphics [trim=0 0 0 0, clip, width=1.0\textwidth, angle = 0]{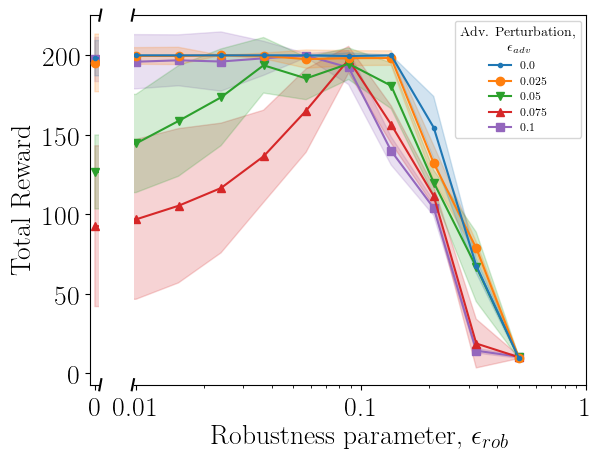}
	\subcaption{Robustness to Adversaries (Cartpole)}\label{fig:carrl_cartpole_adv_curves}
	\end{minipage}
	\begin{minipage}{0.45\linewidth}
	\centering\includegraphics [trim=0 0 0 0, clip, width=1.0\textwidth, angle = 0]{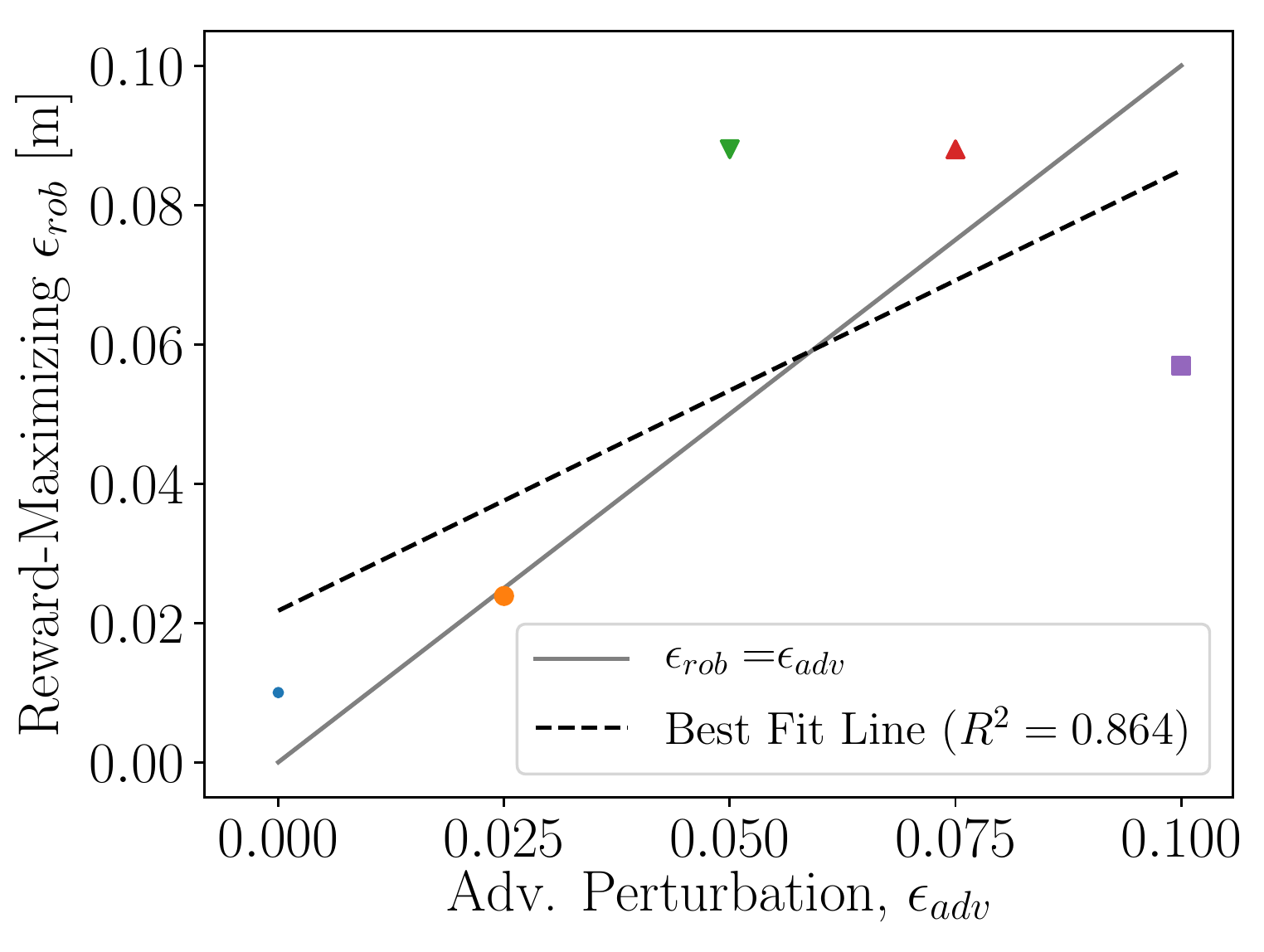}
	\subcaption{Choosing $\epsilon_\text{rob}$ in Presence of Adversary (Cartpole)}\label{fig:carrl_cartpole_adv_best_epsilon}
	\end{minipage}\\
	\begin{minipage}{0.45\linewidth}
	\centering\includegraphics [trim=0 0 0 0, clip, width=1.0\textwidth, angle = 0]{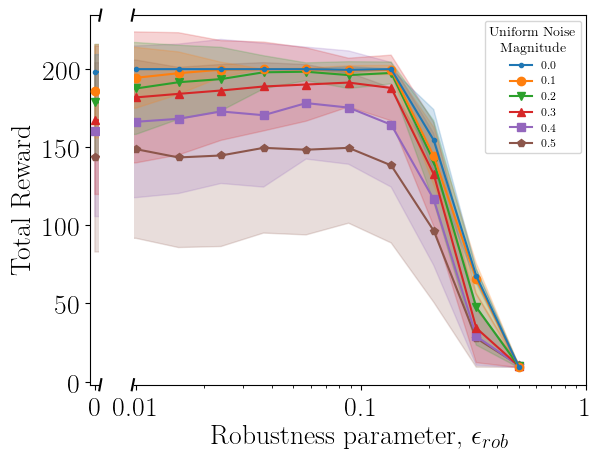}
	\subcaption{Robustness to Sensor Noise (Cartpole)}\label{fig:carrl_cartpole_noise_curves}
	\end{minipage}
	\begin{minipage}{0.45\linewidth}
	\centering\includegraphics [trim=0 0 0 0, clip, width=1.0\textwidth, angle = 0]{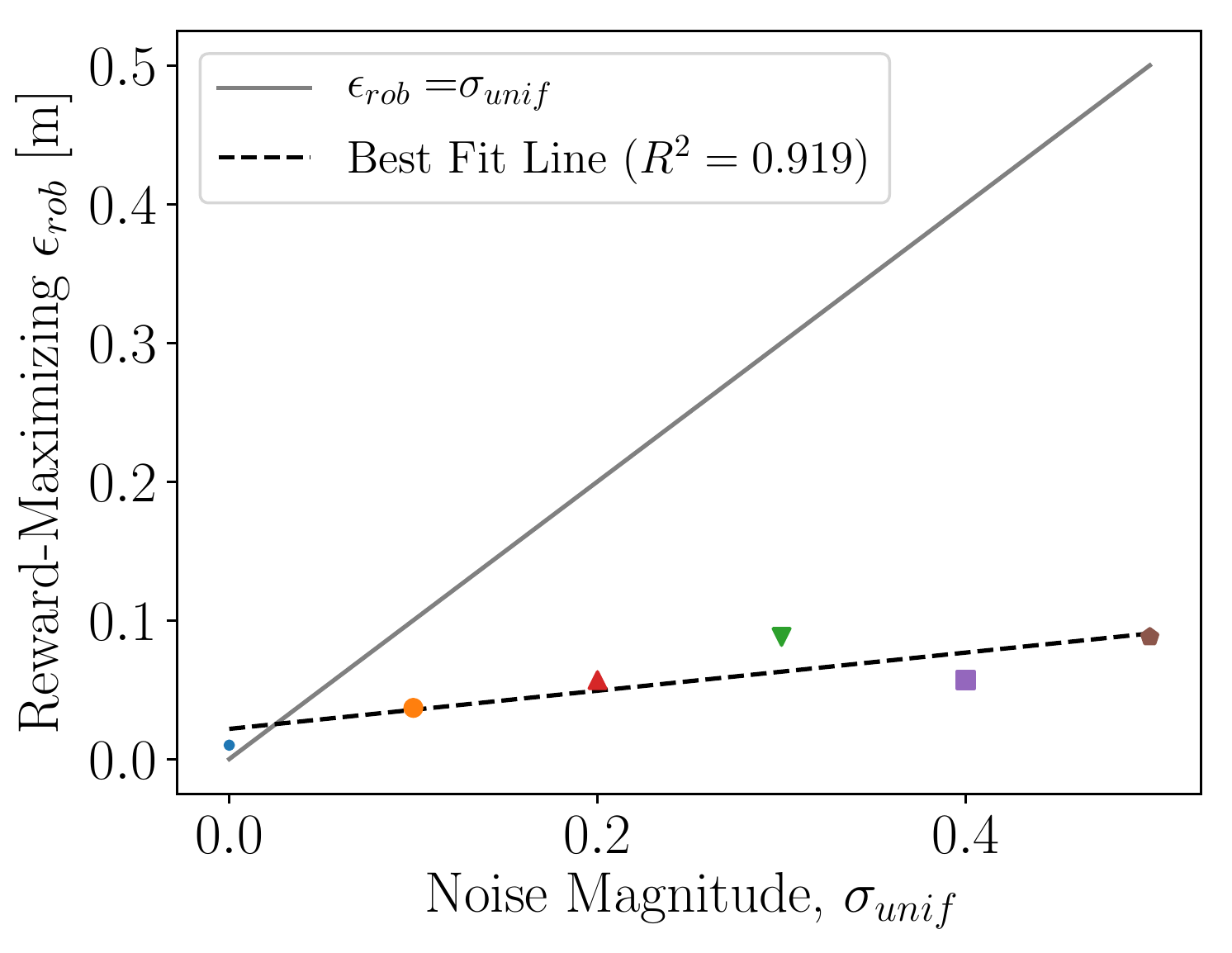}
	\subcaption{Choosing $\epsilon_\text{rob}$ in Presence of Noise (Cartpole)}\label{fig:carrl_cartpole_noise_best_epsilon}
	\end{minipage}
\caption[Results on Cartpole.]{
Results on Cartpole.
\acrshort*{carrl} recovers performance (measured by reward received) by adding robustness under various magnitudes of adversarial (a) and uniform noise (c) perturbations.
Each curve in (a,c) correspond to a different magnitude of perturbation, and $\epsilon_\text{rob}=0$ corresponds to zero robustness, i.e., \acrshort*{dqn}.
For all adversary/noise magnitudes, \acrshort*{carrl} becomes overly conservative for large $\epsilon_\text{rob}$, and the performance degrades.
Thus, choosing the best $\epsilon_\text{rob}$ for a particular perturbation magnitude can be guided by the curves in (b,d).
}
\label{fig:carrl_cartpole}
\end{figure*}

In the cartpole task~\citep{barto1983neuronlike,Brockman_2016}, the reward is the number of time steps (capped at $200$) that a pole remains balanced upright ($\pm12^{\circ}$ from vertical) on a cart that can translate along a horizontal track.
The state vector, ${\bm{s}=\left[p_{cart}, v_{cart}, \theta_{pole}, \dot{\theta}_{pole}\right]^T}$ and action space, ${\bm{a}\in\{\mathrm{push\ cart\ left}, \mathrm{push\ cart\ right}\}}$ are defined in~\cite{Brockman_2016}.
A $2$-layer, $4$-unit network was trained in an environment without any observation perturbations using an open-source \acrshort*{dqn} implementation~\cite{stable-baselines} with Bayesian Optimization used to find training hyperparameters.
The trained \acrshort*{dqn} is evaluated against perturbations of various magnitudes, shown in~\cref{fig:carrl_cartpole}.
In this domain, robustness and perturbations are applied to all states equally, i.e., ${\bm{\epsilon}_\text{adv} = \epsilon_\text{adv} \cdot \mathbbm{1}\in\mathbbm{R}^4}$, ${\bm{\epsilon}_\text{rob} = \epsilon_\text{rob} \cdot \mathbbm{1}\in\mathbbm{R}^4}$ and ${\bm{\sigma}_{unif} = \sigma_{unif} \cdot \mathbbm{1}\in\mathbbm{R}^4}$.

Each curve in~\cref{fig:carrl_cartpole_adv_curves,fig:carrl_cartpole_noise_curves} corresponds to the average reward received under different magnitudes of adversarial, or uniform noise perturbations, respectively.
The reward of a nominal \acrshort*{dqn} agent drops from $200$ under perfect measurements to $105$ under adversarial perturbation of magnitude $\epsilon_\text{adv}=0.075$ (blue and red reward at x-axis, ${\epsilon_\text{rob}=0}$ in \cref{fig:carrl_cartpole_adv_curves}) or to $145$ under uniform noise of magnitude $\sigma_{\text{unif}}=0.5$ (\cref{fig:carrl_cartpole_noise_curves}).
For $\epsilon_\text{rob}>0$ (moving right along x-axis), the \acrshort*{carrl} algorithm considers an increasingly large range of states in its worst-case outcome calculation.
Accordingly, the algorithm is able to recover some of the performance lost due to imperfect observations.
For example, with $\epsilon_\text{adv}=0.075$ (red triangles), \acrshort*{carrl} achieves 200 reward using \acrshort*{carrl} with $\epsilon_\text{rob}=0.1$.
The ability to recover performance is due to \acrshort*{carrl} selecting actions that consider the worst-case state (e.g., a state in which the pole is closest to falling), rather than fully trusting the perturbed observations.

However, there is again tradeoff between robustness and conservatism.
For large values of $\epsilon_\text{rob}$, the average reward declines steeply for all magnitudes of adversaries and noise, because \acrshort*{carrl} considers an excessively large set of worst-case states (more detail provided in~\cref{sec:results:intuition_on_bounds}).

The reward-maximizing choice of $\epsilon_\text{rob}$ for a particular class and magnitude of perturbations is explored in~\cref{fig:carrl_cartpole_adv_best_epsilon,fig:carrl_cartpole_noise_best_epsilon}.
Similarly to the collision avoidance domain, the best choice of $\epsilon_\text{rob}$ is close to $\epsilon_\text{adv}$ under adversarial perturbations and less than $\sigma_{\text{unif}}$ under uniform noise perturbations.

These results use 200 random seeds causing different initial conditions.

\subsection{Atari Pong Domain}\label{sec:results:pong}

\begin{figure*}
	\centering
	\centering\includegraphics[page=1,trim=0 150 90 150, clip, width=1.0\textwidth, angle = 0]{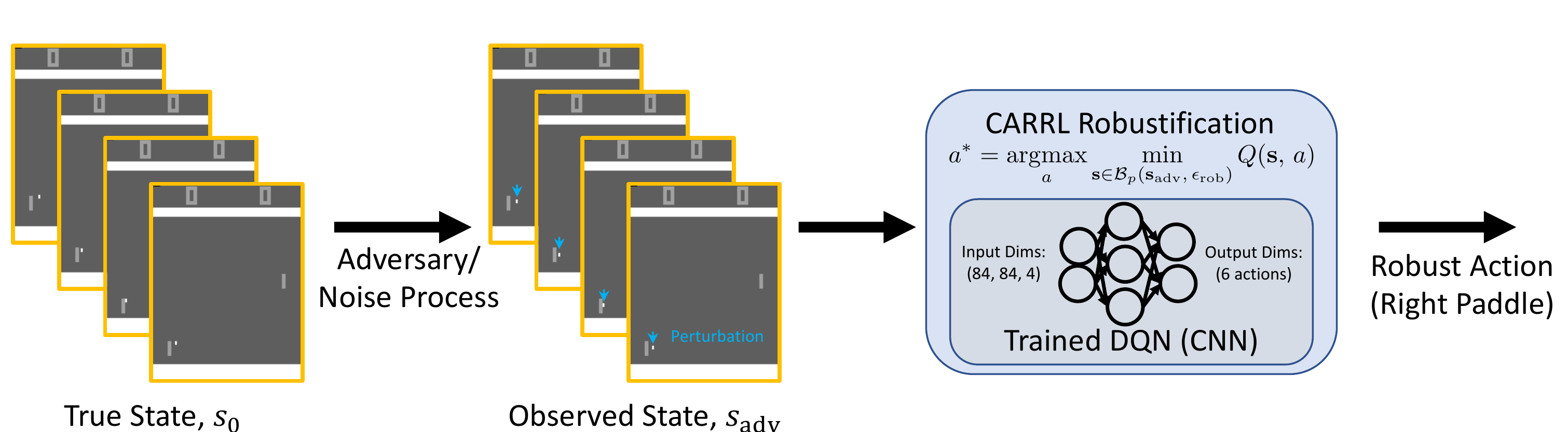}
\caption[Adversary \& Defense in Atari Pong.]{

Adversary \& Defense in Atari Pong.
At each timestep, given the true state (the past 4 $210 \times 160$ grayscale image frames), the adversary finds the pixels corresponding to the ball and moves them downward by $\epsilon_\text{adv}$ pixels.
The CARRL algorithm enumerates the possible un-perturbed states, according to $\epsilon_\text{rob}$, re-shapes the observation to $(84, 84, 4)$, and makes a DQN forward pass to compute $Q_L$ from~\cref{eq:robust_optimal_action_rule}.
The resulting robust-optimal action is implemented, which moves the environment to a new true state, and the cycle repeats.
}
\label{fig:carrl_pong_arch}
\end{figure*}

\begin{figure*}
	\centering
	\begin{subfigure}{0.5\textwidth}
        	\centering
\includegraphics [trim=0 0 0 0, clip, height=2.5in, angle = 0]{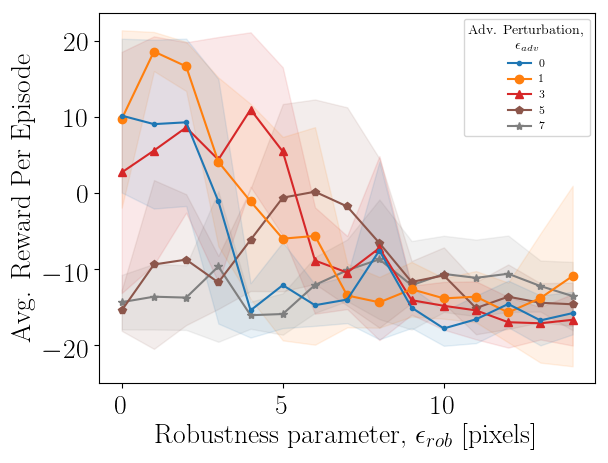}
        \caption{Robustness to Adversaries (Atari Pong)}
        \label{fig:carrl_pong_curves_all}
    \end{subfigure}%
    \begin{subfigure}{0.5\textwidth}
	\centering
        \includegraphics [trim=0 0 0 0, clip, height=2.5in, angle = 0]{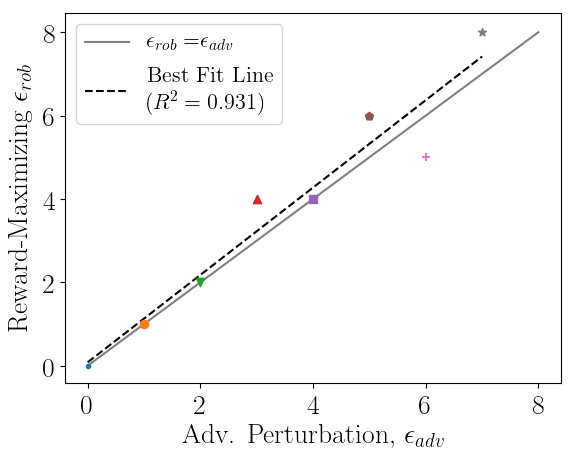}
        \caption{Choosing $\epsilon_\text{rob}$ in Presence of Adversary (Atari Pong)}
        \label{fig:carrl_pong_curves_best}
    \end{subfigure}
\caption[Results on Atari Pong.]{
Results on Atari Pong.
\acrshort*{carrl} recovers performance (measured by reward received) by adding robustness under various magnitudes pixel perturbations.
In (a), each curve corresponds to a different magnitude of perturbation, and $\epsilon_\text{rob}=0$ corresponds to zero robustness, i.e., \acrshort*{dqn}.
For all adversary/noise magnitudes, \acrshort*{carrl} recovers performance roughly until $\epsilon_\text{rob}=\epsilon_\text{adv}$, and then \acrshort*{carrl} becomes overly conservative for large $\epsilon_\text{rob}$, and the performance degrades.
To tune $\epsilon_\text{rob}$, (b) shows the reward-maximizing value of $\epsilon_\text{rob}$ is closely correlated with the (unknown) adversary's magnitude, $\epsilon_\text{adv}$.
Both (a, b) demonstrate similar behavior to that observed in collision avoidance and cartpole tasks, but on a high-dimensional task, with input vector of $(84, 84, 4)$ images and a more complicated DQN architecture (multi-layer CNN from~\cite{Mnih_2015}).
}
\label{fig:carrl_pong_curves}
\end{figure*}

To evaluate CARRL on a task of high dimension, we implemented an adversary and robust defense in image space for the Atari Pong environment~\cite{bellemare13arcade,mott1995stella}.
The framework from~\cite{stable-baselines,rl-zoo} enabled training a DQN on the standard \texttt{PongNoFrameskip-v4} task.
In this task, the observation vector has dimension $(84, 84, 4)$, which represents the past 4 screen frames as grayscale images, and agent chooses from 6 discrete actions to move the right paddle up/down (4 of the actions have no effect).
In addition to a higher-dimensional input vector than in the prior domains (collision avoidance and cartpole), the DQN architecture~\cite{Mnih_2015} used for Pong, consisting of 3 convolutional layers, 2 fully connected layers, all with ReLU activations, is substantially more complicated than the DQNs used above.

\cref{fig:carrl_pong_arch} shows the adversary/defense architecture used in this work's Pong experiments.
The true state is shown on the left, as frames with dimension $(210, 160, 4)$.
An adversary finds the ball in each frame by filtering based on pixel intensity and then shifts the pixels corresponding to the ball downward by the allowed magnitude, $\epsilon_\text{adv}$.
The perturbation is limited to keep the ball within the playable area when the ball is near the perimeter.

This perturbation, if not accounted for, causes a nominal DQN agent act as if the ball is lower than it really is, which causes the agent to position its paddle too low and thus miss the ball and lose a point.
Intuitively, given knowledge of the adversary's magnitude allowance, the robust behavior should be to position the paddle in the middle of all possible ball positions to ensure the paddle at least strikes the ball.

Recall that the previous domains had continuous state spaces, which motivated the calculation of a lower bound, $Q_l$, on the worst-case Q-value.
Because the Pong state space is discrete (pixels have integer values $\in [0, 255]$), the calculation of $Q_L$ can be done exactly by enumerating possible states and querying the DQN for all possibilities in parallel.
Thus, CARRL returns the robust-optimal action, $a^{*}$, from~\cref{eq:robust_optimal_action_rule} in this domain.
CARRL enumerates all possible states by shifting the ball position upward by each integer value $\in [0, \epsilon_\text{rob}]$, leading to $\epsilon_\text{rob}+1$ components of the DQN forward pass.
Note that the adjusted perturbed observation is then downscaled into the $(84, 84, 4)$ vector for the DQN architecture, where the ball-moving operations are done in the raw $(210, 160, 4)$ images to avoid the complexities of blurring/interpolation when finding the ball pixels.

After designing the adversary/defense for this domain, we ran 50 games with various combinations of $\epsilon_\text{rob}$ and $\epsilon_\text{adv}$, as shown in~\cref{fig:carrl_pong_curves}.
In Pong, the two players repeatedly play until one player reaches 21 points, and the reward returned by the environment at the end of an episode is the difference in the two players' scores (e.g., lose $21-6$ $\Rightarrow$ $-15$ reward, and $|R|\leq 21$).

\cref{fig:carrl_pong_curves_all} shows similar behavior as in the other domains.
When $\epsilon_\text{rob}=0$ (left-most points), the average reward decreases as the adversary's strength increases (from 0 to 7 pixels).
Then, as $\epsilon_\text{rob}$ is increased, the reward generally increases, then reaches a peak around $\epsilon_\text{rob}\approx\epsilon_\text{adv}$, then decreases because the agent is too conservative.
Furthermore, \cref{fig:carrl_pong_curves_best} shows that the reward-maximizing choice of $\epsilon_\text{rob}$ is correlated with the adversary's strength, $\epsilon_\text{adv}$.
These results suggest that the proposed framework for robustification of learned policies can also be deployed in high-dimensional tasks, such as Pong, with high-dimensional CNNs, such as the DQN architecture from~\cite{Mnih_2015}.

\subsection{Computational Efficiency}\label{sec:results:computation}

For the cartpole task, one forward pass with bound calculation takes on average $0.68\pm 0.06$ms, which compares to a forward pass of the same \acrshort*{dqn} (nominal, without calculating bounds) of $0.24\pm 0.03$ms; for collision avoidance, it takes $1.85 \pm 1.62$ms (\acrshort*{carrl}) and $0.30 \pm 0.04$ms (\acrshort*{dqn}), all on one i7-6700K CPU.
In our implementation, the bound of all actions (i.e., 2 or 11 discrete actions for the cartpole and collision avoidance domain, respectively) are computed in parallel.
While the \acrshort*{dqn}s used in this work are relatively small, \citep{Weng_2018} shows that the runtime of Fast-Lin scales linearly with the network size, and a recent GPU implementation offers faster performance~\citep{gpu_implementation_crown}, suggesting \acrshort*{carrl} could be implemented in real-time for higher-dimensional \acrshort*{rl} tasks, as well.

\subsection{Robustness to Behavioral Adversaries}\label{sec:results:behavioral_adversary}

\begin{figure*}
	\centering
	\begin{minipage}{0.33\linewidth}
	\centering\includegraphics [trim=0 0 0 0, clip, width=1.0\textwidth, angle = 0]{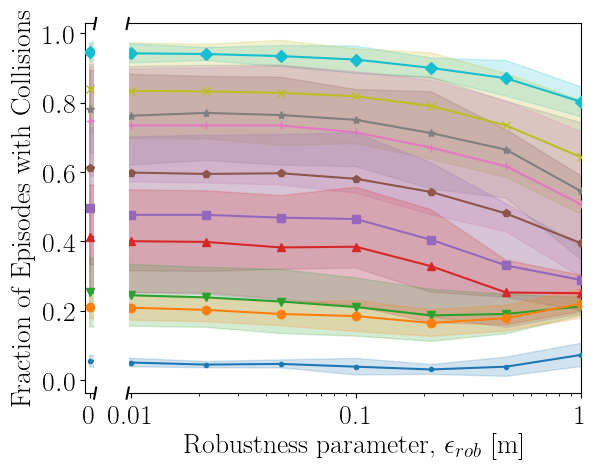}
	\subcaption{}\label{fig:collab_collisions}
	\end{minipage}%
	\begin{minipage}{0.33\linewidth}
	\centering\includegraphics [trim=0 0 0 0, clip, width=1.0\textwidth, angle = 0]{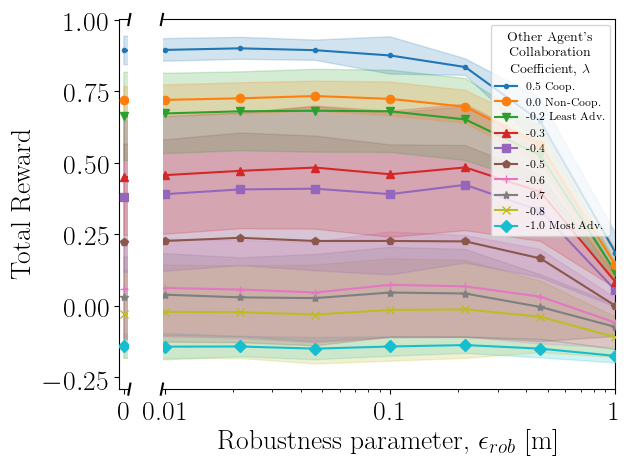}
	\subcaption{}\label{fig:collab_reward}
	\end{minipage}%
	\begin{minipage}{0.33\linewidth}
	\centering\includegraphics [trim=0 0 0 0, clip, width=1.0\textwidth, angle = 0]{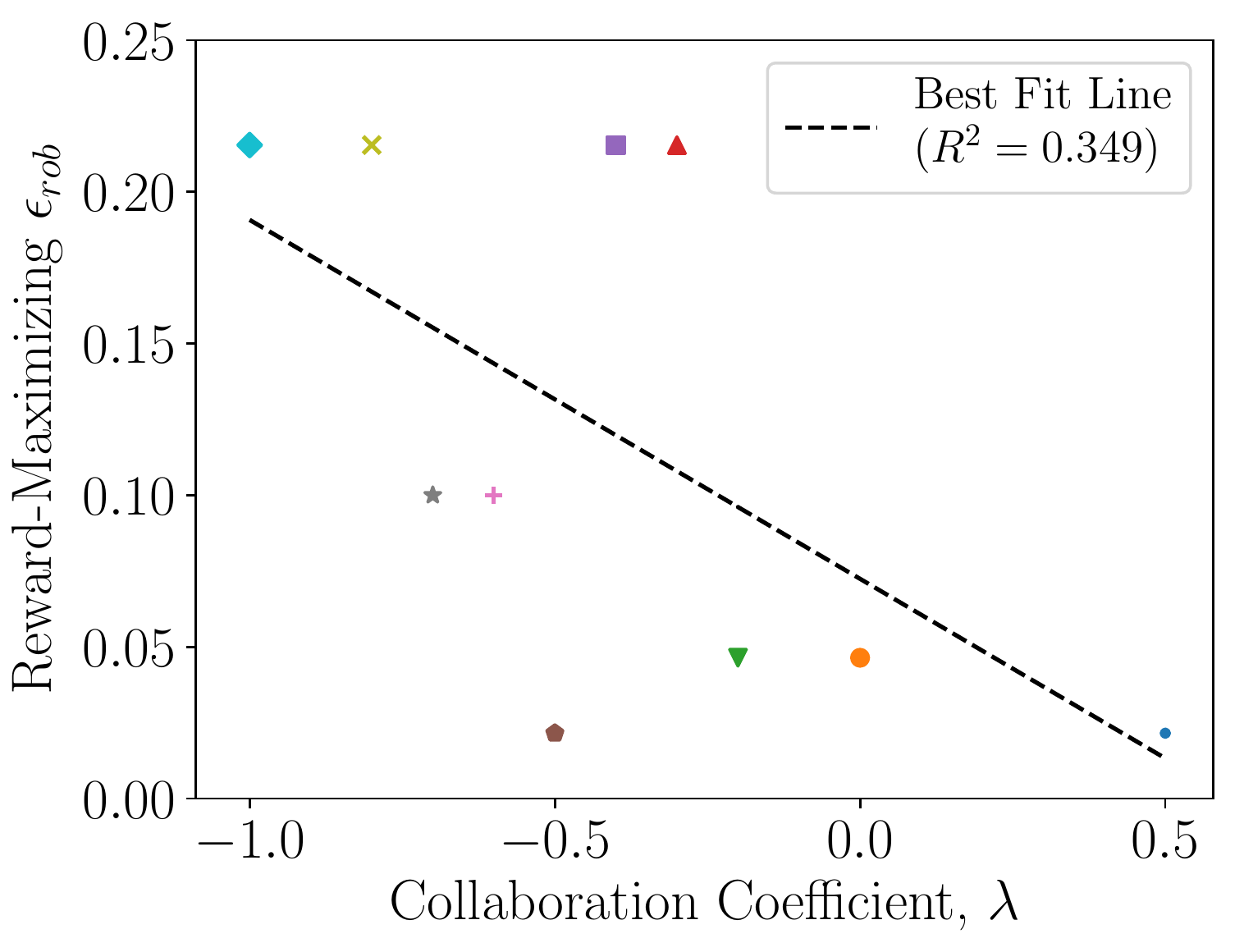}
	\subcaption{}\label{fig:collab_eps_rob_maximizer}
	\end{minipage}%
\caption[Robustness to Adversarial Behavior]{
Robustness to Adversarial Behavior.
Each curve shows a different magnitude of adversarial \textit{behavior} of another agent in the collision avoidance task.
The adversarially behaving other agents (negative collaboration coefficient) are able to cause many collisions with a \acrshort*{carrl} agent trained among cooperative and non-cooperative agents.
Although \acrshort*{carrl} was not explicitly designed to handle behavioral adversaries, \acrshort*{carrl} can reduce the number of collisions by providing robustness in the other agent's position measurement.
CARRL's effect on reward (b) is not as strong as in~\cref{fig:carrl_collision_avoidance}, but the reward-maximizing ${\epsilon_\text{rob}>0}$ against all adversaries (c), and there is a trend of larger $\epsilon_\text{rob}$ working well against stronger adversaries.
}
\label{fig:collab_vs_eps}
\end{figure*}

In addition to observational perturbations, many real-world domains also require interaction with other agents, whose \textit{behavior} could be adversarial.
In the collision avoidance domain, this can be modeled by an environment agent who actively tries to collide with the \acrshort*{carrl} agent, as opposed to the various cooperative or neutral behavior models seen in the training environment described earlier.
Although the \acrshort*{carrl} formulation does not explicitly consider behavioral adversaries, one can introduce robustness to this class of adversarial perturbation through robustness in the observation space, namely by specifying uncertainty in the other agent's position.
In other words, requiring the \acrshort*{carrl} agent to consider worst-case positions of another agent while selecting actions causes the \acrshort*{carrl} agent to maintain a larger spacing, which in turn prevents the adversarially behaving agent from getting close enough to cause collisions.

In the collision avoidance domain, we parameterize the adversarial ``strength'' by a collaboration coefficient, $\lambda$, where $\lambda=0.5$ corresponds to a nominal \acrshort*{orca} agent (that does half the collision avoidance), $\lambda=0$ corresponds to a non-cooperative agent that goes straight toward its goal, and $\lambda\in[-1,0)$ corresponds to an adversarially behaving agent.
Adversarially behaving agents sample from a Bernoulli distribution (every 1 second) with parameter $\lvert\lambda\rvert$. 
If the outcome is 1, the adversarial agent chooses actions directly aiming into the \acrshort*{carrl} agent's projected future position, otherwise, it chooses actions moving straight toward its goal position.
Thus, $\lambda=-1$ means the adversarial agent is always trying to collide with the \acrshort*{carrl} agent.

The idea of using observational robustness to protect against behavioral uncertainty is quantified in~\cref{fig:collab_vs_eps}, where each curve corresponds to a different behavioral adversary.
Increasing the magnitude of $\lambda<0$ (increasingly strong adversaries) causes collisions with increasing frequency, since the environment transition model is increasingly different from what was seen during training.
Increasing \acrshort*{carrl}'s robustness parameter $\epsilon_\text{rob}$ leads to a reduction in the number of collisions, as seen in~\cref{fig:collab_collisions}.
Accordingly, the reward received increases for certain values of $\epsilon_\text{rob}>0$ (seen strongest in red, violet curves; note y-axis scale is wider than in~\cref{fig:carrl_collision_avoidance_adv_rew}).
Although the impact on the reward function is not as large as in the observational uncertainty case, \cref{fig:collab_eps_rob_maximizer} shows that the reward-maximizing choice of $\epsilon_\text{rob}$ has some negative correlation the adversary's strength (i.e., the defense should get stronger against a stronger adversary).

The same trade-off of over-conservatism for large $\epsilon_\text{rob}$ exists in the behavioral adversary setting.
Because there are perfect observations in this experiment (just like training), \acrshort*{carrl} has minimal effect when the other agent is cooperative (blue, ${\lambda=0.5}$).
The non-cooperative curve (orange, ${\lambda=0}$) matches what was seen in~\cref{fig:carrl_collision_avoidance} with $\epsilon_\text{adv}=0$ or $\sigma_{\text{unif}}=0$.
100 test cases with 5 seeds were used.

\subsection{Comparison to LP Bounds}\label{sec:results:carrl_vs_lp}

\begin{figure*}
	\centering
	\begin{minipage}{0.33\linewidth}
	\centering\includegraphics [trim=0 0 0 0, clip, width=1.0\textwidth, angle = 0]{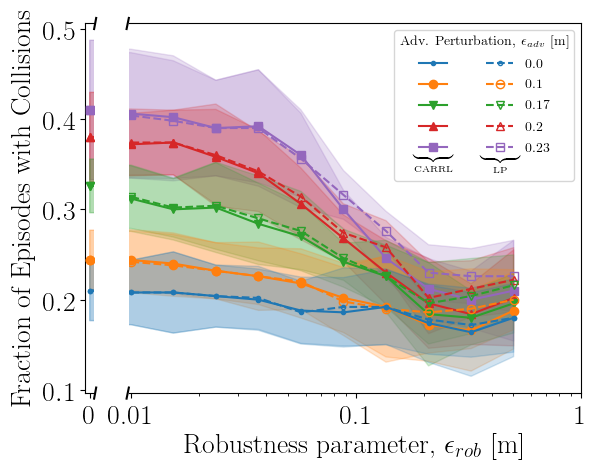}
	\subcaption{}\label{fig:carrl_vs_lp_collisions}
	\end{minipage}%
	\begin{minipage}{0.33\linewidth}
	\centering\includegraphics [trim=0 0 0 0, clip, width=1.0\textwidth, angle = 0]{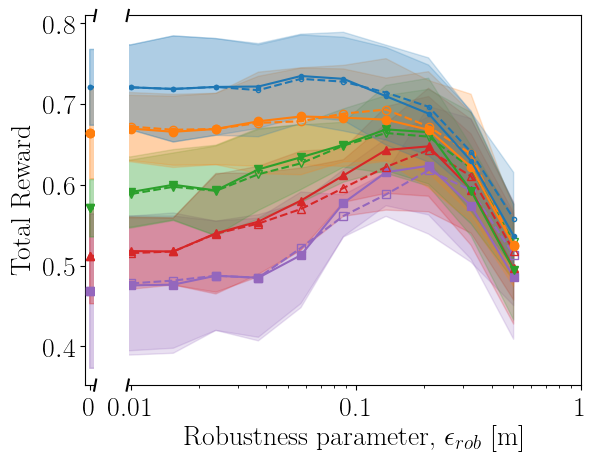}
	\subcaption{}\label{fig:carrl_vs_lp_reward}
	\end{minipage}%
	\begin{minipage}{0.33\linewidth}
	\centering\includegraphics[trim=0 0 0 0, clip, width=1.0\linewidth, angle = 0]{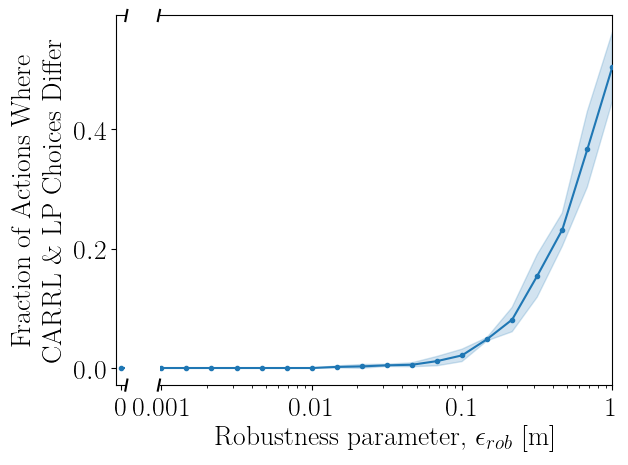}
	\subcaption{}\label{fig:carrl_vs_lp_actions}
	\end{minipage}%
\caption[Greedy Convex Bounds vs.\ \acrshort*{lp}]{Greedy Convex Bounds vs. \acrshort*{lp}. Using efficient but greedy bounds on the adversarial polytope, \acrshort*{carrl}'s action-selection produces similar performance to that of action-selection using the less relaxed \acrshort*{lp} (cf. \textbf{LP-ALL} in~\citep{Weng_2018b}) (solid vs. dashed curves in each color). In (c), \acrshort*{carrl} and \acrshort*{lp} select identical actions on $>99\%$ of timesteps (across 60 episodes) for small $\epsilon_\text{rob}$, but the two methods diverge as the greedy bounds become overly conservative. Thus, for small-to-moderate $\epsilon_\text{rob}$, the extra computation required for the \acrshort*{lp} does not have substantial effect on decision-making.}
\label{fig:carrl_vs_lp}
\end{figure*}

As described in~\cref{sec:related_work}, the convex relaxation approaches (e.g., Fast-Lin) provide relatively loose, but fast-to-compute bounds on \acrshort*{dnn} outputs.
\Cref{eq:Q_geq_QL_geq_QLP_geq_Ql} relates various bound tightnesses theoretically, which raises the question: how much better would the performance of an \acrshort*{rl} agent be, given more computation time to better approximate the worst-case outcome, $Q_{L}$?

In~\cref{fig:carrl_vs_lp}, we compare the performance of an agent following the \acrshort*{carrl} decision rule, versus one that approximates $Q_L(\bm{s}_\text{adv},\bm{a})$ with the full (non-greedy) primal convex relaxed \acrshort*{lp} (in the collision avoidance domain with observational perturbations).
Our implementation is based on the \textbf{LP-ALL} implementation provided in~\citep{Weng_2018b}.
The key takeaway is that there is very little difference: the \acrshort*{carrl} curves are solid and the \acrshort*{lp} curves are dashed, for various settings of adversarial perturbation, $\epsilon_\text{adv}$.
The small difference in the two algorithms is explained by the fact that \acrshort*{carrl} provides extra conservatism versus the \acrshort*{lp} (\acrshort*{carrl} accounts for a worse worst-case than the \acrshort*{lp}), so the \acrshort*{carrl} algorithm performs slightly better when $\epsilon_\text{rob}$ is set too small, and slightly worse when $\epsilon_\text{rob}$ is too large for the particular observational adversary.

This result is further explained by~\cref{fig:carrl_vs_lp_actions}, where it is shown that \acrshort*{carrl} and \acrshort*{lp}-based decision rules choose the same action ${>99\%}$ of the time for ${\epsilon<0.1}$, meaning in this experiment, the extra time spent computing the tighter bounds has little impact on the \acrshort*{rl} agent's decisions.

\subsection{Intuition on Certified Bounds}\label{sec:results:intuition_on_bounds}

\begin{figure*}
	\begin{minipage}{0.49\linewidth}
	\centering\includegraphics [trim=0 0 0 0, clip, width=1.0\textwidth, angle = 0]{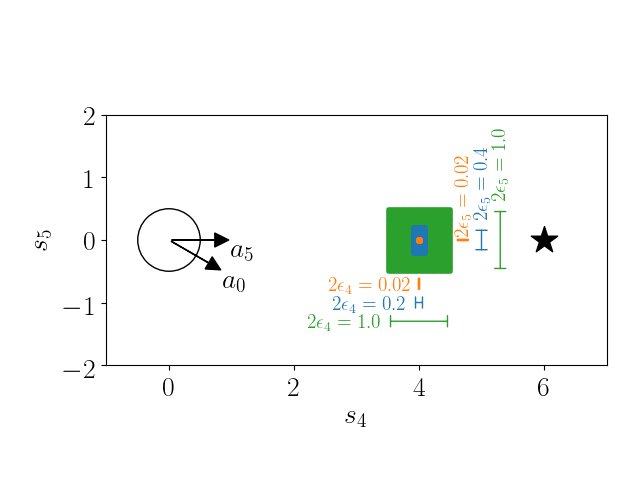}
	\subcaption{Position Uncertainty of Another Agent [m]}\label{fig:visualize_q_state_space_0}
	\end{minipage}
	\begin{minipage}{0.49\linewidth}
	\centering\includegraphics [trim=0 0 0 0, clip, width=1.0\textwidth, angle = 0]{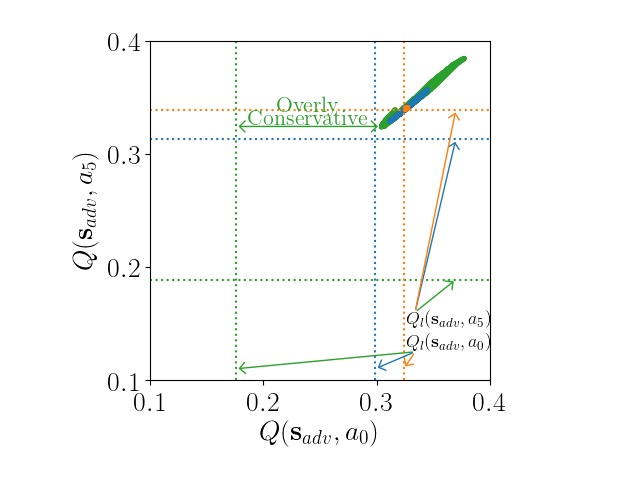}
	\subcaption{Corresponding Certified Bounds \& Sampled DQN Outputs}\label{fig:visualize_q_action_space_0}
	\end{minipage}
	\caption[Influence of $\bm{\epsilon}_\text{rob}$ on Q-Values]{
	Influence of $\bm{\epsilon}_\text{rob}$ on Q-Values.
	In (a), the \acrshort*{carrl} agent ($\bigcirc$) has a goal ($\bigstar$) at (6,0) and decides between two actions, $a_0$ and $a_5$.
	A second agent could be centered somewhere in the colored regions, corresponding to different values of $\bm{\epsilon}_\text{rob}$.
	In (b) are the corresponding Q-values for those possible states (orange, blue green regions in top-right), exhaustively sampled in each $\bm{\epsilon}_\text{rob}$-Ball.
	As $\bm{\epsilon}_\text{rob}$ increases, the spread of possible Q-values increases.
	\acrshort*{carrl}'s lower bounds on each action, $Q_l(\bm{s}_\text{adv}, a_{0})$, $Q_l(\bm{s}_\text{adv}, a_{5})$, are depicted by the dotted lines.
	Conservatism is measured by the gap between the dashed line and the left/bottom-most sampled point.
	For small $\bm{\epsilon}_\text{rob}$ (orange), the bounds are tight; for moderate $\bm{\epsilon}_\text{rob}$ (blue) are moderately conservative, and for large $\bm{\epsilon}_\text{rob}$ (green), the linear approximation of \acrshort*{relu} degrades, causing excessive conservatism.
	}
	\label{fig:visualize_q0}
	\begin{minipage}{0.49\linewidth}
	\centering\includegraphics [trim=0 0 0 0, clip, width=1.0\textwidth, angle = 0]{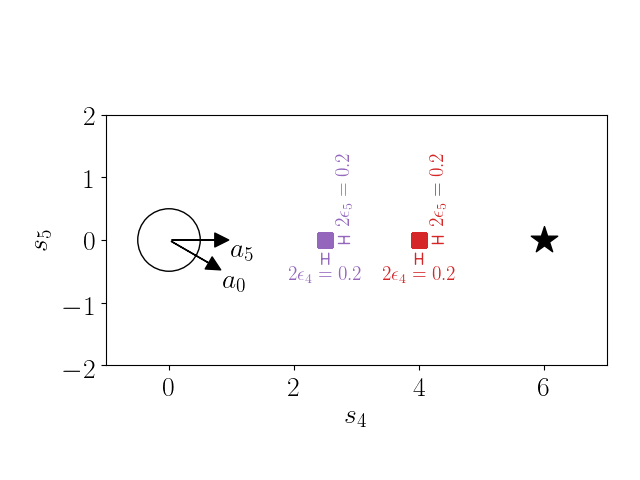}
	\subcaption{Position Uncertainty of Another Agent [m]}\label{fig:visualize_q_state_space_1}
	\end{minipage}
	\begin{minipage}{0.49\linewidth}
	\centering\includegraphics [trim=0 0 0 0, clip, width=1.0\textwidth, angle = 0]{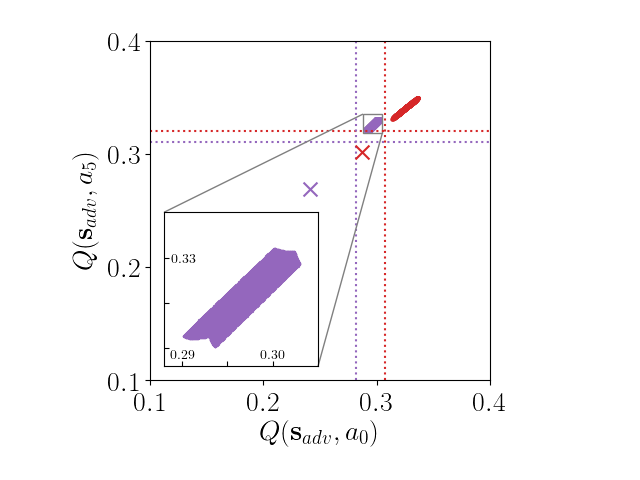}
	\subcaption{Corresponding Certified Bounds \& Sampled DQN Outputs}\label{fig:visualize_q_action_space_1}
	\end{minipage}
	\caption[Influence of $\bm{s}_\text{adv}$ on Q-Values.]{
	Influence of $\bm{s}_\text{adv}$ on Q-Values.
	For the same $\bm{\epsilon}_\text{rob}$, the spread of Q-values are shown for two examples of $\bm{s}_\text{adv}$.
	When the other agent is close (purple), the Q-values are lower than when the other agent is far (red).
	A closer look at one of the non-convex adversarial polytopes is inset in (b).
	Moreover, if one instead used the heuristic of simply inflating the other agent's radius, the Q-values would lie at the $\times$'s -- in both cases, radius inflation is more conservative (further toward bottom-left) than \acrshort*{carrl}.
	}
\label{fig:visualize_q1}
\end{figure*}

Visual inspection of actual adversarial polytopes and the corresponding efficiently computed bounds provides additional intuition into the \acrshort*{carrl} algorithm.
The state uncertainty's mapping into an adversarial polytope in the Q-value space is visualized in~\cref{fig:visualize_q0,fig:visualize_q1}.
In~\cref{fig:visualize_q_state_space_0}, a \acrshort*{carrl} agent observes another agent (of the same size) positioned somewhere in the concentric, colored regions.
The state uncertainty, drawn for various values of $\bm{\epsilon}_\text{rob}$ with $\ell_{\infty}$ norm manifests itself in~\cref{fig:visualize_q_action_space_0} as a region of possible $Q(\bm{s}_\text{adv}, a_j)$ values, for each action, $a_j$.
Because there are only two dimensions of state uncertainty, we can exhaustively sample $Q$-values for these states.
To visualize the corresponding Q-value region in 2D, consider just two actions, $a_0, a_5$ (right-most and straight actions, respectively).

The efficiently computed lower bounds, $Q_l$ for each action are shown as dotted lines for each of the $\bm{\epsilon}_\text{rob}$-Balls.
Note that for small $\bm{\epsilon}_\text{rob}$ (orange) the bounds are quite tight to the region of possible Q-values.
A larger $\bm{\epsilon}_\text{rob}$ region (blue) leads to looser (but still useful) bounds, and this case also demonstrates $\bm{\epsilon}_\text{rob}$ with non-uniform components ($\epsilon_4=0.1, \epsilon_5=0.2$, where $(4,5)$ are the indices of the other agent's position in the state vector)
For large $\bm{\epsilon}_\text{rob}$ (green), the bounds are very loose, as the lowest sampled Q-value for $a_0$ is $0.3$, but $Q_l(\bm{s}_\text{adv},a_0)=0.16$.

This increase in conservatism with $\bm{\epsilon}_\text{rob}$ is explained by the formulation of Fast-Lin, in that linear bounds around an ``undecided'' ReLU become looser for larger input uncertainties.
That is, as defined in~\cref{eq:sigma_lk_uk}, the lower linear bound of an undecided ReLU in layer $k$, neuron $i$ is $\frac{u^{(k)}_i \cdot l^{(k)}_i}{u^{(k)}_i + l^{(k)}_i}<0$ for $z^{(k)}_i=l^{(k)}_i$, whereas a ReLU can never output a negative number.
This under-approximation gets more extreme for large $\epsilon_{rob,i}$, since $u^{(k)}_i, l^{(k)}_i$ become further apart in the input layer, and this conservatism is propagated through the rest of the network.
Per the discussion in~\citep{Weng_2018b}, using the same slope for the upper and lower bounds has computational advantages but can lead to conservatism.
The \acrshort*{crown}~\citep{Weng_2018b} algorithm generalizes this notion and allows adaptation in the slopes; connecting \acrshort*{crown} with \acrshort*{carrl} is a minor change left for future work.

Moreover, expanding the possible uncertainty magnitudes without excessive conservatism and while maintaining computational efficiency could be an area of future research.
Nonetheless, the lower bounds in \acrshort*{carrl} were sufficiently tight to be beneficial across many scenarios.

In addition to $\bm{\epsilon}_\text{rob}$, \cref{fig:visualize_q_state_space_1} considers the impact of variations in $\bm{s}_\text{adv}$: when the other agent is closer to the \acrshort*{carrl} agent (purple), the corresponding Q-values in~\cref{fig:visualize_q_action_space_1} are lower for each action, than when the other agent is further away (red).
Note that the shape of the adversarial polytope (region of Q-value) can be quite complicated due to the high dimensionality of the \acrshort*{dnn} (inset of~\cref{fig:visualize_q_action_space_1}).
Furthermore, the $\times$ symbols correspond to the Q-value if the agent simply inflated the other agent's radius to account for the whole region of uncertainty.
This heuristic is highly conservative (and domain-specific), as the $\times$'s have lower Q-values than the efficiently computed lower bounds for each action in these examples.


\section{Future Directions} \label{sec:future}

In laying a foundation for certified adversarial robustness in deep RL, this work offers numerous future research directions in connecting deep RL algorithms and real-world, safety-critical systems.

For example, extensions of \acrshort{carrl} to other \acrshort{rl} tasks will raise the question: how does the robustness analysis extend to continuous action spaces?
When $\bm{\epsilon}_\text{adv}$ is unknown, as in our empirical results, how could $\bm{\epsilon}_\text{rob}$ be tuned efficiently online while maintaining robustness guarantees?
How can the online robustness estimates account for uncertainties from the training process (e.g., unexplored states) in a guaranteed manner?

The key factor in the performance of robustness algorithms is the ability to precisely describe the uncertainty over which to be robust.
This work provides the $\bm{\epsilon}_\text{rob}$, $p_\text{rob}$ hyperparameters to describe various shapes and sizes of uncertainties in different dimensions of the observation space and shows how this description could be applied under non-uniform, probabilistic uncertainties or a model of behavioral uncertainty.
However, real-world systems also include other types of environmental uncertainties.
For instance, observational uncertainties beyond $L_p$ balls could be studied for certifiably robust defenses in deep RL (e.g., \cite{brown1712adversarial,wong2019wasserstein} in supervised learning).
Moreover, how can one protect against an adversary that is allowed to plan $n$ timesteps into the future (e.g., extending~\cite{wang2019verification} to model-free RL)?
Are there certifiably robust defenses to uncertainties in the state transition model (e.g., disturbance forces)?

Lastly, we assume in~\cref{sec:approach:optimal_cost_function} that the training process has access to unperturbed observations and causes the network to converge to the optimal value function.
In reality, the DQN will not exactly converge to Q*, and this additional source of uncertainty should be taken into account when computing bounds on Q.
For example, a simple model of this uncertainty is to assume that training causes the DQN outputs to converge to within $\alpha$ of the true Q-values, i.e., $\lvert Q_\text{DQN}(\mathbf{s},\,a) - Q^*(\mathbf{s},\,a) \rvert \leq \alpha$.
Then, the upper/lower bounds on DQN outputs can be inflated by $\alpha$ to reflect true Q-values, i.e., $Q_L(\mathbf{s},\,a) \leq Q_\text{DQN}(\mathbf{s},\,a) \leq Q_U(\mathbf{s},\,a) \implies Q_L(\mathbf{s},\,a)-\alpha \leq Q^*(\mathbf{s},\,a) \leq Q_U(\mathbf{s},\,a)+\alpha$.
However, this adjustment will have no impact on the action-selection rule in~\cref{eq:opt_cost_fn}.
Moreover, ensuring convergence properties hold remains a major open problem in deep learning.

Understanding each of these areas of extension will be crucial in providing both performance and robustness guarantees for deep RL algorithms deployed on real-world systems.

\section{Conclusion} \label{sec:conclusion}
This work adapted deep \acrshort*{rl} algorithms for application in safety-critical domains, by proposing an add-on \textit{certifiably robust} defense to address existing failures under adversarially perturbed observations and sensor noise.
The proposed extension of robustness analysis tools from the verification literature into a deep \acrshort*{rl} formulation enabled efficient calculation of a lower bound on Q-values, given the observation perturbation/uncertainty.
These guaranteed lower bounds were used to efficiently solve a robust optimization formulation for action selection to provide maximum performance under worst-case observation perturbations.
Moreover, the resulting policy comes with a \textit{certificate} of solution quality, even though the true state and optimal action are unknown to the certifier due to the perturbations.
The resulting policy (added onto trained DQN networks) was shown to improve robustness to adversaries and sensor noise, causing fewer collisions in a collision avoidance domain and higher reward in cartpole.
Furthermore, the proposed algorithm was demonstrated in the presence of adversaries in the behavior space and image space, compared against a more time-intensive alternative, and visualized for particular scenarios to provide intuition on the algorithm's conservatism.

\appendix

\section*{Certified vs. Verified Terminology}\label{sec:cert_vs_verif}
Now that we have introduced our algorithm, we can more carefully describe what we mean by a \textit{certifiably robust} algorithm.

First, we clarify what it means to be \textit{verified}.
In the context of neural network robustness, a network is \textit{verified robust} for some nominal input if no perturbation within some known set can change the network's decision from its nominal decision.
A \textit{complete} verification algorithm would correctly label the network as either \textit{verified robust} or \textit{verified non-robust} for any nominal input and perturbation set.
In practice, verification algorithms often involve network relaxations, and the resulting algorithms are  \textit{sound} (any time the network gives an answer, the answer is true), but sometimes return that they can not verify either property.

A \textit{certificate} is some piece of information that allows an algorithm to quantify its solution quality.
For example, \citep{boyd2004convex} uses a dual feasible point as a \textit{certificate} to bound the primal solution, which allows one to compute a bound on any feasible point's suboptimality even when the optimal value of the primal problem is unknown.
When an algorithm provides a certificate, the literature commonly refers to the algorithm as \textit{certified}, \textit{certifiable}, or \textit{certifiably \underline{\hspace{1cm}}} (if it provides a \textit{certificate of \underline{\hspace{1cm}}-ness}).

In this work, we use a robust optimization formulation to consider worst-case possibilities on state uncertainty (\cref{eq:robust_optimal_action_rule,eq:opt_cost_fn}); this makes our algorithm \textit{robust}.
Furthermore, we provide a certificate on how sub-optimal the robust action recommended by our algorithm is, with respect to the optimal action at the (unknown) true state (\cref{eqn:carrl:suboptimality_certificate}).
Thus, we describe our algorithm as \textit{certifiably robust} or that it provides \textit{certified adversarial robustness}.

As an aside, some works in computer vision/robot perception propose \textit{certifiably correct} algorithms~\citep{bandeira2016note,Yang20arXiv-TEASER}.
In those settings, there is often a combinatorial problem that \textit{could} be solved optimally with enough computation time, but practitioners prefer an algorithm that efficiently computes a (possibly sub-optimal) solution and comes with a certificate of solution quality.
Our setting differs; the state uncertainty means that recovering the optimal action for the true state is not a matter of computation time.
Thus, we do not aim for \textit{correctness} (choosing the optimal action for the true state), but rather choose an action that maximizes worst-case performance.


%



\section*{Acknowledgment}
This work is supported by Ford Motor Company. The authors greatly thank Tsui-Wei (Lily) Weng for providing code for the Fast-Lin algorithm and insightful discussions, as well as Rose Wang and Dr. Kasra Khosoussi.

\bibliographystyle{IEEEtran}
\bibliography{biblio}


\begin{IEEEbiography}[{\includegraphics[width=1in,height=1.25in,clip,keepaspectratio]{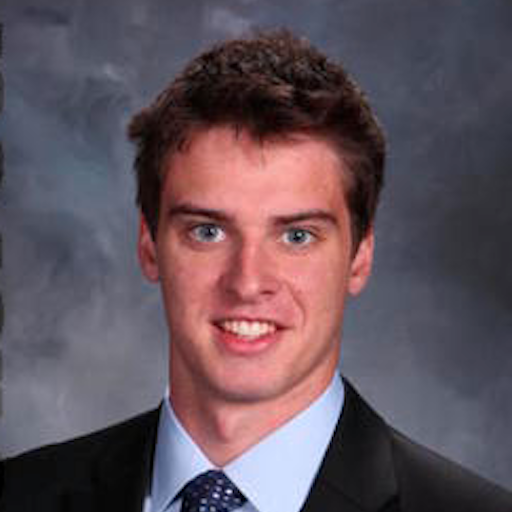}}]{Michael Everett}
received the S.B., S.M., and Ph.D. degrees in mechanical engineering from the Massachusetts Institute of Technology (MIT), in 2015, 2017, and 2020, respectively. He is currently a Postdoctoral Associate with the Department of Aeronautics and Astronautics at MIT. His research lies at the intersection of machine learning, robotics, and control theory, with specific interests in the theory and application of safe and robust neural feedback loops. He was an author of works that won the Best Paper Award on Cognitive Robotics at IROS 2019, the Best Student Paper Award and a Finalist for the Best Paper Award on Cognitive Robotics at IROS 2017, and a Finalist for the Best Multi-Robot Systems Paper Award at ICRA 2017. He has been interviewed live on the air by BBC Radio and his team’s robots were featured by Today Show and the Boston Globe.
\end{IEEEbiography}

\begin{IEEEbiography}[{\includegraphics[width=1in,height=1.25in,clip,trim=5 5 10 5, keepaspectratio]{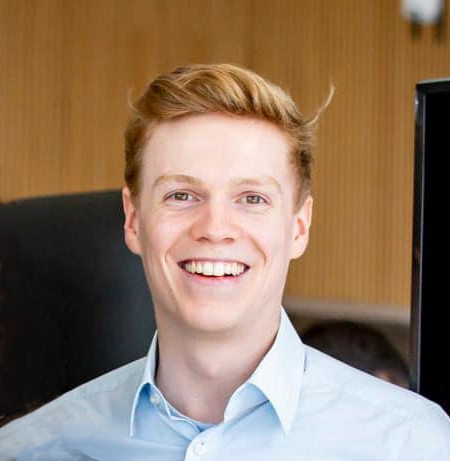}}]{Bj\"{o}rn L\"{u}tjens}
is currently a Ph.D. Candidate in the Human Systems Laboratory of the Department of Aeronautics and Astronautics at MIT. 
He has received the S.M. degree in Aeronautics and Astronautics from MIT in 2019 and the B.Sc. degree in Engineering Science from Technical University of Munich in 2017. 
His research interests include deep reinforcement learning, bayesian deep learning, and climate and ocean modeling.
\end{IEEEbiography}

\begin{IEEEbiography}[{\includegraphics[trim=80 0 50 0,width=.9in,clip,keepaspectratio]{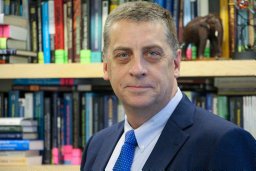}}]{Jonathan P. How} is the Richard C. Maclaurin Professor of Aeronautics and Astronautics at the Massachusetts Institute of Technology.  He received a B.A.Sc. (aerospace) from the University of Toronto in 1987, and his S.M. and Ph.D. in Aeronautics and Astronautics from MIT in 1990 and 1993, respectively, and then studied for 1.5 years at MIT as a postdoctoral associate. Prior to joining MIT in 2000, he was an assistant professor in the Department of Aeronautics and Astronautics at Stanford University.  Dr. How was the editor-in-chief of the IEEE Control Systems Magazine (2015-19) and is an associate editor for the AIAA Journal of Aerospace Information Systems and the IEEE Transactions on Neural Networks and Learning Systems. He was an area chair for International Joint Conference on Artificial Intelligence (2019) and will be the program vice-chair (tutorials) for the Conference on Decision and Control (2021).  He was elected to the Board of Governors of the IEEE Control System Society (CSS) in 2019 and is a member of the IEEE CSS Technical Committee on Aerospace Control and the Technical Committee on Intelligent Control. He is the Director of the Ford-MIT Alliance and was a member of the USAF Scientific Advisory Board (SAB) from 2014-17. His research focuses on robust planning and learning under uncertainty with an emphasis on multiagent systems, and he was the planning and control lead for the MIT DARPA Urban Challenge team.  His work has been recognized with multiple awards, including the 2020 AIAA Intelligent Systems Award. He is a Fellow of IEEE and AIAA.   
\end{IEEEbiography}

\end{document}